\documentclass[lettersize,journal]{IEEEtran}
\usepackage[switch]{lineno}  
\usepackage{amsmath,amsfonts}
\usepackage{amssymb}
\usepackage{times}
\usepackage{algorithm}
\usepackage{algpseudocode}
\usepackage{array}
\usepackage[caption=false,font=normalsize,labelfont=sf,textfont=sf]{subfig}
\usepackage{orcidlink}
\usepackage{textcomp}
\usepackage{stfloats}
\usepackage{url}
\usepackage{verbatim}
\usepackage{graphicx}
\usepackage{cite}
\usepackage{booktabs}
\usepackage{multirow}
\usepackage{threeparttable}
\usepackage{makecell}
\usepackage{tabularx}
\usepackage[table]{xcolor}
\colorlet{blue}{black}
\definecolor{newblue}{RGB}{0,0,255}
\colorlet{newblue}{black}
\usepackage{colortbl}
\usepackage{pifont}
\newlength{\qualheight}
\newcommand{\cmark}{\ding{51}}

\definecolor{oursblue}{RGB}{225,240,255} 
\begin{document}
\title{$\mathcal{B}^{3}$-Net: Controlled Posterior Bridge Learning for\\ Multi-Task Dense Prediction
}

\author{Meihua~Zhou\,\orcidlink{0009-0008-4309-7208} and Li~Yang%
\thanks{This work was supported by the Scientific Research Project of Higher Education Institutions in Anhui Province (2024AH053451) and the Anhui Province 2025 University Science and Engineering Teachers Enterprise Secondment Practice Program (2025jsqygz42).}%
\thanks{Meihua Zhou is with Wannan Medical University, Wuhu, China, and also with the University of Chinese Academy of Sciences, Beijing, China (e-mail: zhoumeihua25@mails.ucas.ac.cn).}%
\thanks{Li Yang is with Wannan Medical University, Wuhu, China (e-mail: yangli@wnmc.edu.cn).}%
\thanks{Corresponding author: Li Yang.}%
}


\markboth{Journal of \LaTeX\ Class Files,~Vol.~14, No.~8, August~2021}%
{Shell \MakeLowercase{\textit{et al.}}: A Sample Article Using IEEEtran.cls for IEEE Journals}

\makeatletter
\renewcommand{\@IEEEpubidpullup}{2.5\baselineskip}
\makeatother

\IEEEpubid{\parbox[b]{\textwidth}{\footnotesize
Copyright \copyright~2026 IEEE. Personal use of this material is permitted.
However, permission to use this material for any other purposes must be
obtained from the IEEE by sending an email to pubs-permissions@ieee.org.}}

\maketitle

\begin{abstract}
Multi-task dense prediction aims to solve complementary pixel-level tasks within a unified model, including semantic segmentation, depth estimation, surface normal estimation, and edge detection. Recent methods have improved cross-task interaction through attention, prompts, routing, diffusion, Mamba, and bridge features. However, most decoder-side interactions still organize task evidence implicitly. They often assume that task evidence can be fused according to similarity or affinity, while ignoring that evidence reliability varies across tasks and spatial locations. This limitation can contaminate the shared representation and amplify negative transfer when unreliable information is redistributed to task branches. In this paper, we propose $\mathcal{B}^{3}$-Net, a controlled posterior bridge learning framework for multi-task dense prediction. The key idea is to decompose decoder-side cross-task interaction into three coupled steps, including reliability estimation, posterior bridge construction, and bounded redistribution. Specifically, the Precision Field Estimator estimates patch-wise evidence precision from task-reference alignment and local variation. The Posterior Bridge Operator constructs a precision-weighted posterior bridge through heteroscedastic evidence fusion, \textcolor{blue}{with source contributions determined by spatial precision.} The Contractive Dispatch Operator redistributes the bridge to each task branch through a bounded update, reducing the risk of uncontrolled feature injection. Extensive experiments on NYUD-v2, PASCAL-Context, and Cityscapes demonstrate that $\mathcal{B}^{3}$-Net achieves competitive or superior overall trade-offs compared with representative CNN-based, Transformer-based, diffusion-based, Mamba-based, and bridge-feature-based methods. Backbone-matched comparisons, operator ablations, mean-bridge analysis, precision-error verification, contraction analysis, task-set analysis, and efficiency evaluation further support the contribution of controlled posterior bridge learning across the evaluated backbones and task configurations. \textcolor{blue}{The code is available at \url{https://github.com/mh-zhou/B3Net}.}
\end{abstract}


\section{Introduction}
\IEEEPARstart{D}{ense} multi-task prediction aims to solve multiple pixel-level scene understanding tasks within a unified model, including semantic segmentation, depth estimation, surface normal estimation, and edge detection. These tasks provide complementary descriptions of the same scene. Semantic segmentation identifies object categories, depth estimation and surface normal estimation characterize geometric structure, and edge detection preserves local discontinuities. By sharing representations across tasks, joint learning can reduce redundant computation and exploit cross-task complementarity, making dense multi-task prediction an important paradigm for efficient visual scene understanding systems~\cite{MTLSurvey,UberNet}.

\begin{figure}[htbp]
\centering
\includegraphics[width=0.35\textwidth]{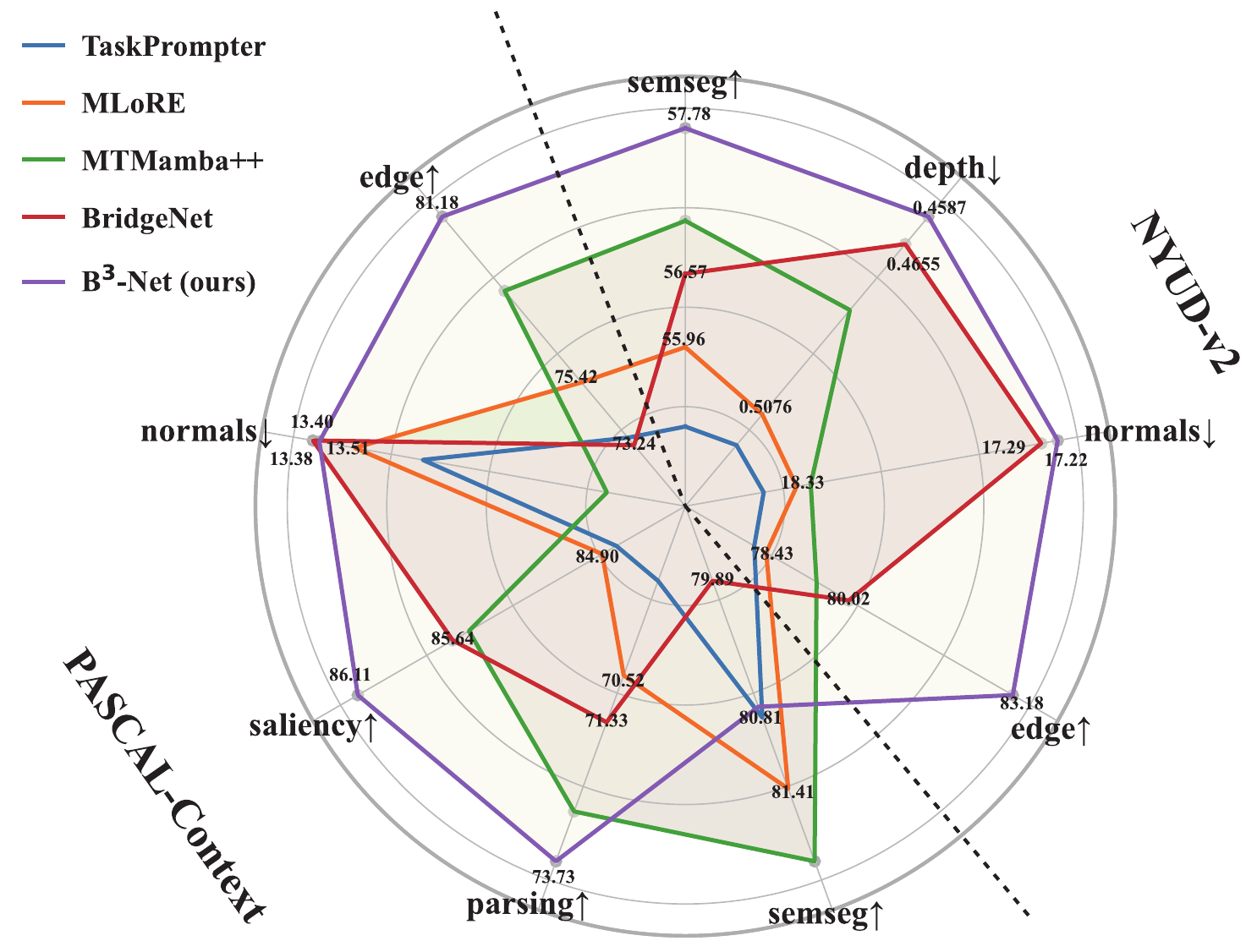}
\caption{\textbf{Overall performance comparison.}
We compare B$^{3}$-Net with representative recent methods on NYUD-v2 and PASCAL-Context using a unified higher-is-better normalization for visualization.}
\label{fig:radar_comparison}
\end{figure}

\IEEEpubidadjcol

Recent methods have shifted from simple parameter sharing to explicit task interaction in the decoder.
{\color{blue}Task-transfer studies identify task compatibility as a key determinant of multi-task benefit beyond task count~\cite{Taskonomy,MTLTransference}.
CNN-based methods transfer information through task-specific decoders, multimodal distillation, layer-wise fusion, and multiscale interaction~\cite{PADNet,NDDRCNN,MTINet,ATRC}.}
Transformer-based methods introduce global spatial and cross-task attention~\cite{InvPT,InvPTPlusPlus,MQTransformer,TSPTransformer}.
{\color{blue}Task prompts jointly model task-generic features, task-specific features, and their interactions~\cite{TaskPrompter}, while adaptive sharing and grouping determine sharing patterns and interacting tasks~\cite{AdaShare,TaskGrouping,AutoLambda}.}
Mixture-of-experts, modular learning, knowledge factorization, diffusion, and Mamba-based decoders improve capacity, specialization, refinement, or long-range modeling efficiency~\cite{MLoRE,ModSquad,FKN,TaskDiffusion,MTMamba,MTMambaPlusPlus}.
Bridge-feature-based methods introduce an explicit intermediate representation to mediate task interaction~\cite{BridgeNet}.

{\color{blue}These designs often leave reliability estimation, shared-state formation, and redistribution implicitly coupled through similarity, affinity, or routing scores.
Evidence quality varies across tasks at boundaries, occlusions, depth discontinuities, and ambiguous semantic regions.
Unreliable evidence can contaminate the shared state and propagate through unrestricted updates, inducing negative transfer.
Multi-objective optimization, adaptive loss weighting, task prioritization, gradient surgery, conflict-averse descent, bargaining-based optimization, and surrogate-based detection address training conflicts primarily at the loss or gradient level~\cite{MGDA,GradNorm,DTP,PCGrad,CAGrad,NashMTL,MTSL}, leaving local evidence selection and decoder redistribution unspecified.
Synergy modeling, self-training, binarized efficient prediction, and uncertainty-weighted robust perception provide further advances~\cite{SEM,MultiTaskSelfTraining,BiMTDP}.}
These complement our goal, which couples reliability-aware evidence organization before shared-state formation with bounded task redistribution.

\textcolor{blue}{To integrate task evidence according to spatially varying reliability and preserve task-specific structures,}
we propose $\mathcal{B}^{3}$-Net, a controlled posterior bridge learning framework for multi-task dense prediction.
The name $\mathcal{B}^{3}$ reflects three coupled principles, \emph{\underline{B}ridge}, \emph{\underline{B}ayesian}, and \emph{\underline{B}ounded}.
\emph{\underline{B}ridge} denotes an explicit shared state for organizing cross-task evidence.
\emph{\underline{B}ayesian} denotes posterior bridge construction under spatially varying precision.
\emph{\underline{B}ounded} denotes contractive bridge-to-task redistribution.
\textcolor{blue}{The Precision Field Estimator, Posterior Bridge Operator, and Contractive Dispatch Operator implement patch-wise precision estimation, precision-weighted posterior inference, and bounded redistribution in a closed loop.}
We evaluate $\mathcal{B}^{3}$-Net on NYUD-v2, PASCAL-Context, and Cityscapes.
Compared with representative CNN-based, Transformer-based, diffusion-based, Mamba-based, and bridge-feature-based decoders, $\mathcal{B}^{3}$-Net achieves stronger overall task trade-offs, as shown in Fig.~\ref{fig:radar_comparison}.
\textcolor{blue}{Backbone-matched comparisons, operator and task-set ablations, and mean-bridge, precision-error, contraction, and efficiency analyses support controlled posterior bridge learning across the evaluated backbones and task configurations.}

The main contributions are summarized as follows.
\begin{itemize}
    \item We identify uncontrolled cross-task information organization as a key source of unstable trade-off in dense multi-task prediction, and formulate decoder interaction as reliability estimation, posterior bridge construction, and controlled redistribution.
    \item We propose a Precision Field Estimator to estimate patch-wise evidence precision \textcolor{blue}{and} derive a Posterior Bridge Operator from heteroscedastic evidence fusion, \textcolor{blue}{coupling local reliability estimation with precision-weighted posterior bridge construction.}
    \item We design a Contractive Dispatch Operator for bounded bridge-to-task redistribution, and validate $\mathcal{B}^{3}$-Net through extensive experiments and mechanism analyses.
\end{itemize}

\section{Related Work}

\begin{figure}[htbp]
\centering
\subfloat[Interaction evolution.]{
    \includegraphics[width=0.85\columnwidth]{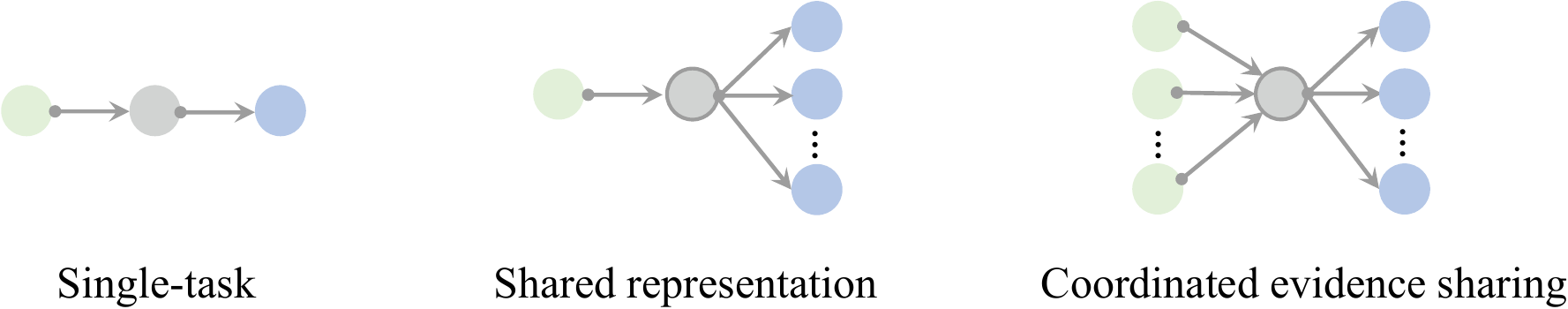}
    \label{fig:motivation_evolution}
}
\\[-2pt]
\subfloat[Interaction topology.]{
    \includegraphics[width=0.85\columnwidth]{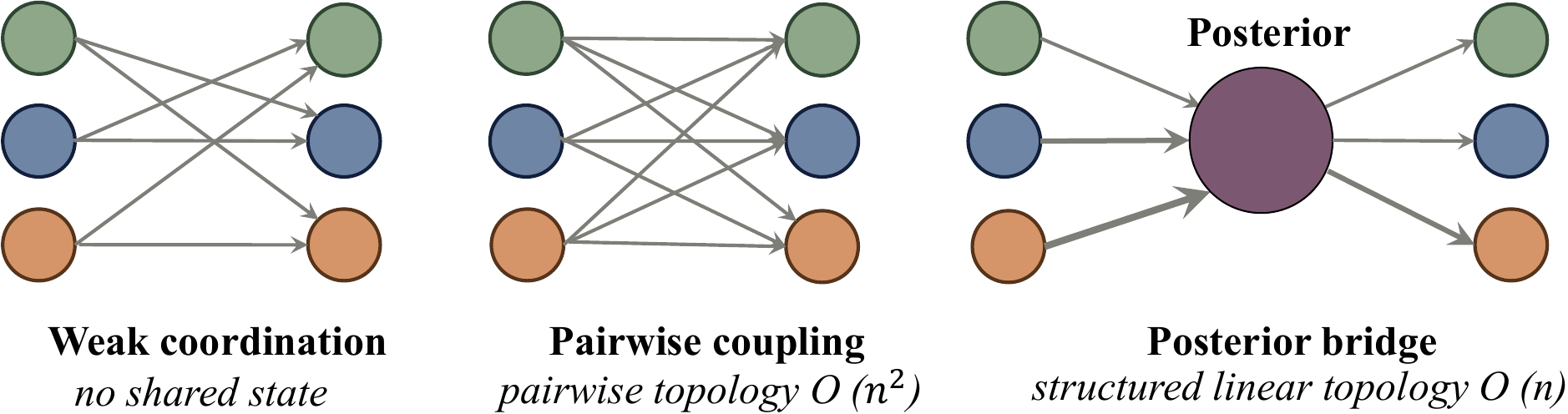}
    \label{fig:motivation_topology}
}
\caption{\textbf{Motivation of structured cross-task interaction.}
Dense multi-task prediction has evolved from weak task coordination to dense pairwise interaction and bridge-mediated organization. Existing designs improve task communication, but the reliability of task evidence and the stability of information redistribution remain under-specified.}
\label{fig:motivation}
\vspace{0.5pt}
\end{figure}
\subsection{Multi-Task Learning for Dense Scene Understanding}
Multi-task learning has been widely explored in visual understanding, including saliency estimation, visual tracking, autonomous driving perception, and dense prediction~\cite{MTRankSaliency,MTCOTracking,UniSparseBEV}. Dense scene understanding focuses on pixel level tasks such as semantic segmentation, depth estimation, surface normal estimation, saliency estimation, and edge detection (see Fig.~\ref{fig:motivation}). Early multi-task methods mainly exploit shared representations and task-specific branches to improve efficiency and encourage task complementarity. Later methods increasingly move the interaction to the decoder, where task-specific representations are more discriminative for dense prediction. CNN based decoders transfer auxiliary information through multimodal distillation, task affinity, pattern propagation, and multiscale interaction~\cite{PADNet,MTINet,PAP,PSD,ATRC}. Transformer based decoders further introduce global spatial and task reasoning through attention, task queries, and cross scale interaction~\cite{InvPT,InvPTPlusPlus,MQTransformer,TSPTransformer}. Prompt based methods encode task generic features, task-specific features, and task relations with learnable task prompts~\cite{TaskPrompter}. Expert based methods improve task specialization through routing and low rank experts~\cite{MLoRE}, while state space models improve long range modeling efficiency~\cite{MTMambaPlusPlus}. Recent adaptive message passing further studies spatially varying task transfer~\cite{ATMPNet}. Bridge based methods introduce an intermediate representation to reduce pair wise transfer and improve the quality of interaction participants~\cite{BridgeNet}. These studies show a clear evolution from parameter sharing to decoder-side information organization. However, most designs still rely on attention affinity, routing probability, message passing, or deterministic feature fusion to organize task evidence. The local reliability of task evidence is usually left implicit before a shared representation is formed.

\begin{figure*}[t]
\centering
\includegraphics[width=0.85\textwidth]{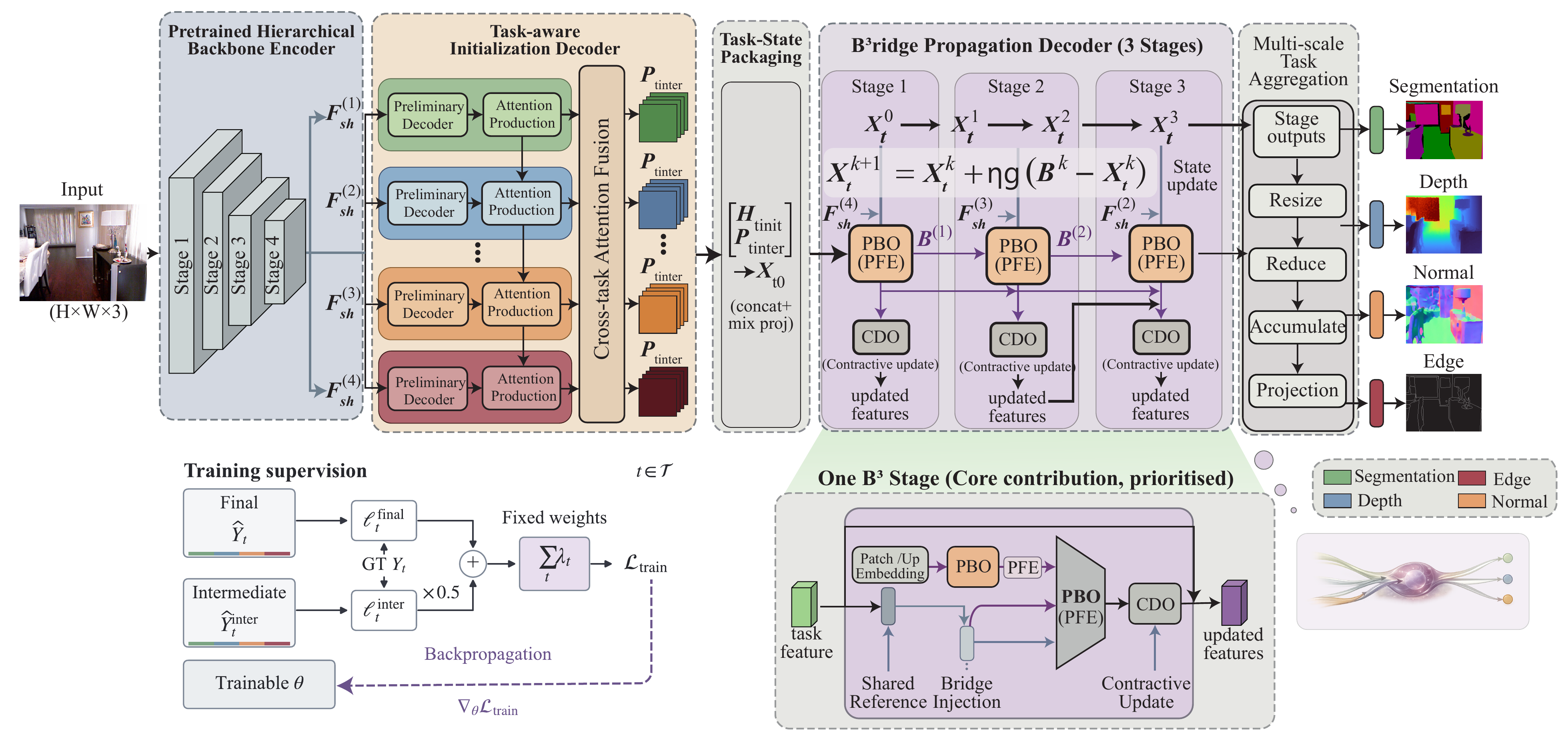}
\setlength{\abovecaptionskip}{-0.2cm} 
\caption{\textbf{Overview of B$^{3}$-Net.}
The network consists of a pretrained hierarchical backbone encoder, a task-aware initialization decoder, a task-state packaging module, a B$^{3}$ bridge propagation decoder, and a multi-scale task aggregation head. The initialization decoder produces preliminary task states from multi-scale shared features. The B$^{3}$ propagation decoder then performs structured cross-task interaction through repeated B$^{3}$ stages. In each stage, PBO and PFE construct a precision-guided posterior bridge from task-specific evidence, while CDO redistributes the bridge state to task branches through a contractive update. The final aggregation head combines multi-scale stage outputs and predicts semantic segmentation, depth, surface normals, and edges. {\color{blue}Final and intermediate losses use fixed task weights $\lambda_t$, with intermediate supervision scaled by $0.5$.
The dashed arrow denotes backpropagation to trainable parameters $\theta$.}}
\label{fig:method_overview}
\vspace{-3pt}
\end{figure*}

\subsection{Reliable Fusion and Propagation}
Reliability aware modeling provides a useful basis for organizing heterogeneous evidence. In computer vision, uncertainty has been used to characterize prediction confidence and to balance multi-task losses for scene geometry and semantics~\cite{KendallUncertainty,KendallMTLUncertainty}. These methods show that uncertain signals should contribute less to learning, but most of them act at the prediction level, task level, or loss level. Dense decoder interaction requires a more local view because task evidence can be reliable in one region and unreliable in another. Bayesian estimation gives a principled formulation for this case. When multiple Gaussian observations have different variances, the posterior estimate is weighted by precision~\cite{BishopPRML,MurphyGaussian}. This provides a statistical foundation for precision-weighted shared-state estimation from heterogeneous task evidence. After aggregation, the shared state also needs stable propagation. Unrestricted residual update or feature injection can amplify unreliable information and disturb task-specific representations. Contraction mapping, Lipschitz bounded transformations, and fixed point models provide a theoretical language for stable iterative refinement~\cite{BanachFixedPoint,LipschitzNN,DEQ}. These results suggest that reliable decoder interaction should consider both precision-weighted evidence aggregation and bounded information propagation. Existing dense multi-task decoders have rarely coupled these two aspects in a unified cross task interaction mechanism.

\section{Method}
\subsection{Method Overview}

\begin{figure*}[t]
\centering
\includegraphics[width=0.85\textwidth]{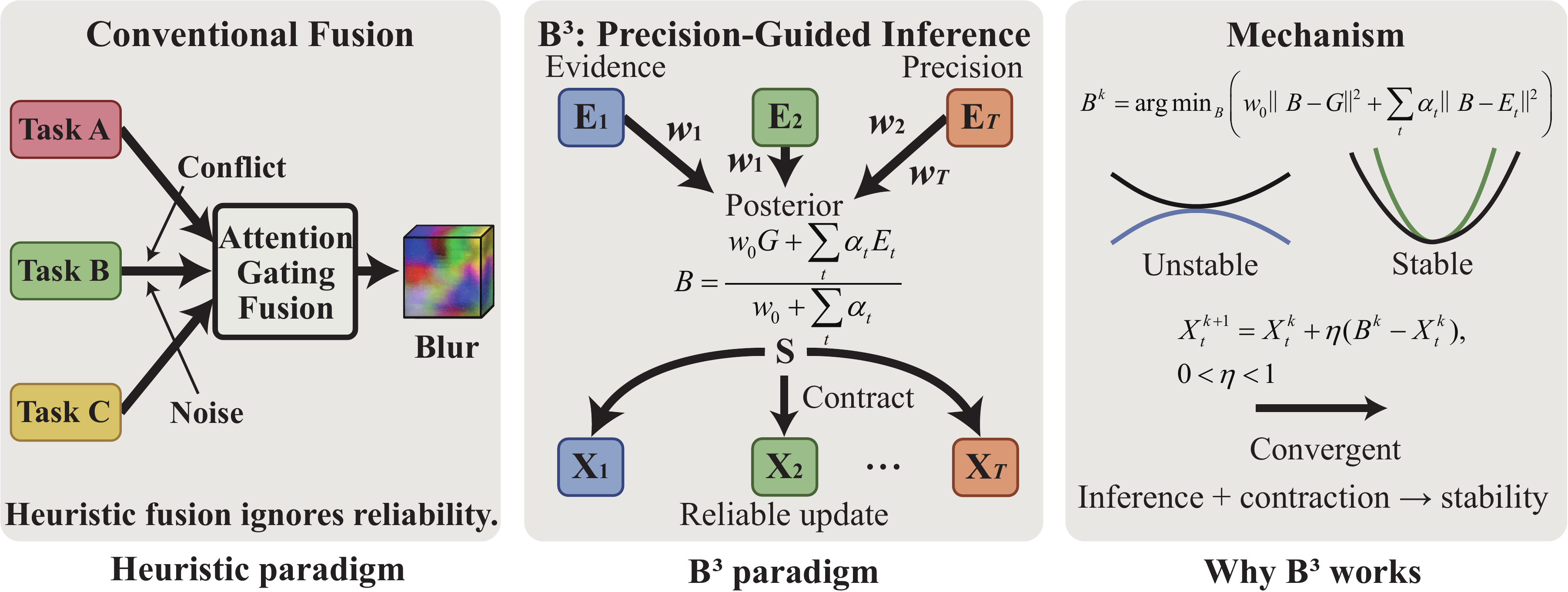}
\caption{\textbf{Conceptual motivation of B$^{3}$-Net.}
Conventional fusion may mix task evidence without considering spatially varying reliability. B$^{3}$-Net constructs a precision-guided posterior bridge from task evidence $\{E_t\}_{t=1}^{T}$ and precision weights $\{\alpha_t\}_{t=1}^{T}$, then redistributes it through a bounded contractive update. The displayed update
$X_t^{k+1}=X_t^k+\eta(B^k-X_t^k)$
\textcolor{blue}{illustrates convergence toward a fixed bridge
with a constant step $0<\eta<1$, while CDO uses
task-specific gated updates.}}
\label{fig:b3_concept}
\end{figure*}

Fig.~\ref{fig:method_overview} shows the overall architecture of $\mathcal{B}^{3}$-Net. Given an input image $x$ and a task set $\mathcal{T}=\{1,\dots,T\}$, where $T$ is the number of tasks, a pretrained hierarchical backbone extracts multi-scale shared features $\{F_{sh}^{(s)}\}_{s=1}^{S}$. A task-aware initialization decoder first produces preliminary task states. These states are then packaged into task representations $\{X_t^{(0)}\}_{t=1}^{T}$ and passed to the B$^{3}$ bridge propagation decoder. At stage $k$, the decoder maintains a task state $X_t^{(k)}$ for each task and a shared bridge $B^{(k)}$ that mediates cross-task interaction. The final multi-scale aggregation head combines stage outputs and predicts all dense tasks.  

{\color{blue} For the three-stage hierarchical decoder, the shared reference is \[ G^{(k)}=\mathrm{RefEmbed}_{k}\!\left(F_{sh}^{(5-k)}\right), \qquad k=1,2,3. \] Here $\mathrm{RefEmbed}_{k}$ comprises the scale-specific feature embedding, a $1\times1$ channel projection, normalization and ReLU, followed by reference normalization. The three stages therefore use backbone levels 4, 3 and 2. Cross-attentive extraction uses $G^{(k)}$ as query and the task state as key and value. \par}

The central challenge in dense MTL is negative transfer. While tasks often exhibit correlation, they are not uniformly consistent. \textcolor{blue}{We focus on the decoder-side propagation of locally inconsistent task evidence.} At each spatial location, task evidence can have different reliability. Some tasks provide stable signals, while others remain uncertain or misleading. Existing feature mixing methods, such as mean aggregation or attention pooling, often assume that task evidence can be fused without explicitly modeling this local reliability. Once a contaminated shared representation is written back through residual addition or overwrite, unreliable information may propagate across task branches and suppress task-specific structures.

We address this problem by casting cross-task interaction as posterior inference under spatially heterogeneous uncertainty, as illustrated in Fig.~\ref{fig:b3_concept}. Fig.~\ref{fig:b3_concept} illustrates bounded bridge-to-task propagation through $X_t^{k+1}=X_t^k+\eta(B^k-X_t^k)$. A task-specific gate modulates this step in the implementation described in Sec.~\ref{sec:cdo}. \textcolor{blue}{Accordingly,} $\mathcal{B}^{3}$-Net estimates a shared latent bridge conditioned on both task evidence and reliability, and injects it back through a bounded update. 

\textcolor{blue}{The three operators are} Precision Field Estimation (PFE) (see Fig.~\ref{fig:pfe_operator}), Posterior Bridge Operator (PBO) (see Fig.~\ref{fig:pbo}), and Contractive Dispatch Operator (CDO) (see Fig.~\ref{fig:cdo_operator}). PFE estimates patch-wise precision, PBO constructs the posterior bridge, and CDO redistributes the bridge to task branches through a bounded update. 

\begin{algorithm}[t]
\caption{B$^{3}$-Net with posterior bridge propagation}
\label{alg:b3_overall}
\small
\begin{algorithmic}[1]
\Require Input image $x$, task set $\mathcal{T}$
\Ensure Predictions $\{\hat{Y}_t\}$
\State Extract shared features $\{F_{sh}^{(s)}\}$
\State Initialize task states\textcolor{blue}{.}
\[
\{H_t^{init}, P_t^{inter}\} \gets \mathrm{InitDecoder}(F_{sh}^{(4)})
\]
\For{each $t$}
\[
X_t^{(0)} \gets \mathrm{MixProj}([H_t^{init}, P_t^{inter}])
\]
\EndFor
\For{$k=1,2,3$}
    \State \textcolor{blue}{$G^{(k)} \gets \mathrm{RefEmbed}_{k}(F_{sh}^{(5-k)})$}
    \For{each $t$}
        \State $E_t^{(k)} \gets \mathrm{Extract}(X_t^{(k-1)}, G^{(k)})$
        \State $\alpha_t^{(k)} \gets \mathrm{PFE}(E_t^{(k)}, G^{(k)})$
    \EndFor
    \State $B^{(k)} \gets \mathrm{PBO}(G^{(k)}, \{E_t^{(k)}\}, \{\alpha_t^{(k)}\})$
    \State \textcolor{blue}{Form $\{ref_t^{(k)}\}$ by learned correction and reference refinement}
    \For{each $t$}
        \State $\textcolor{blue}{X_t^+} \gets \mathrm{CDO}(X_t^{(k-1)},\textcolor{blue}{ref_t^{(k)}},\alpha_t^{(k)})$
        \State \textcolor{blue}{$X_t^{(k)} \gets \mathrm{Refine}_{t,k}(X_t^{(k-1)},X_t^+,ref_t^{(k)})$}
    \EndFor
\EndFor
\State Aggregate $\{X_t^{(k)}\}$ and predict $\hat{Y}_t$
\end{algorithmic}
\end{algorithm}

Algorithm~\ref{alg:b3_overall} presents the stage-level computation, with task-state embeddings and repeated refinement summarized within each stage.
\textcolor{blue}{Within each refinement block, CDO injection precedes reference concatenation and residual convolutional refinement,
represented by $\mathrm{Refine}_{t,k}$.
Training combines final and intermediate supervision as shown in Fig.~\ref{fig:method_overview},
\[
\mathcal L_{\mathrm{train}}=
\sum_{t\in\mathcal T}\lambda_t
\left[\ell_t(\hat Y_t,Y_t)+0.5\,\ell_t(P_t^{inter},Y_t)\right].
\]
Both predictions are aligned to the supervision resolution.
The task weights $\lambda_t$ remain fixed during training and may differ across tasks.
The total loss is backpropagated through the trainable backbone, decoder and prediction heads,
including the precision and gate parameters, without dynamic gradient balancing.}

\subsection{Posterior Bridge Operator}

We first redefine the shared state. \textcolor{blue}{Uniform averaging assigns equal coefficients to evidence with spatially varying quality. PBO uses local precision to determine each source's contribution.}

\begin{figure*}[t]
\centering
\includegraphics[width=0.85\textwidth]{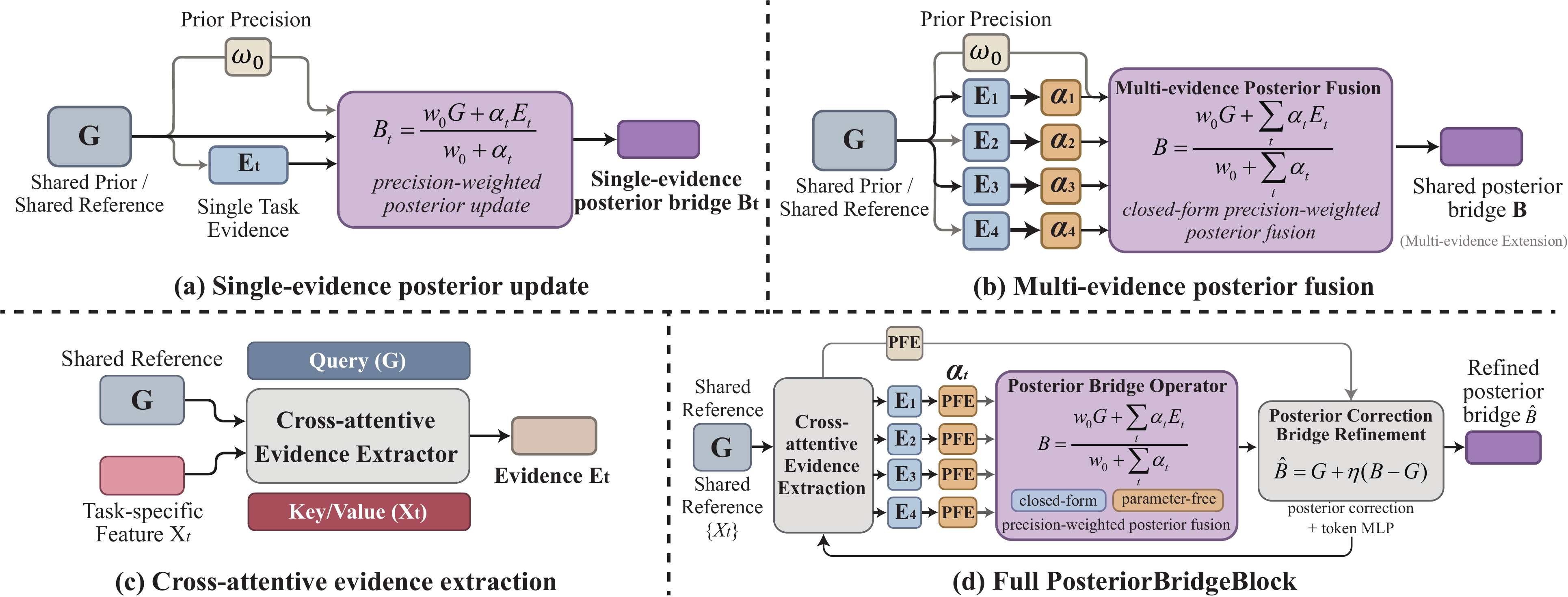}
\vspace{-2pt}
\caption{\textbf{Illustration of PBO.}
(a) A single task evidence $E_t$ is fused with the shared prior/reference $G$ to form a single-evidence posterior bridge through precision-weighted posterior update.
(b) This update is generalized to multi-evidence posterior fusion, where multiple task evidences $\{E_t\}$ are aggregated into a shared posterior bridge $B$ in closed form.
(c) Task evidence is extracted by a cross-attentive evidence extractor, which uses the shared reference as query and the task-specific feature as key/value.
(d) The complete Posterior Bridge Block combines cross-attentive evidence extraction, precision estimation, closed-form posterior fusion, and bounded posterior correction, thereby constructing a reliability-aware bridge reference for subsequent cross-task interaction.}
\label{fig:pbo}
\vspace{-4pt}
\end{figure*}

We \textcolor{blue}{define} the shared state as a posterior latent variable. Let $E_t$ denote task evidence and $G$ denote a shared reference. We define the bridge posterior as
\begin{equation}
p(B \mid G, \{E_t\}) \propto p(B \mid G)\prod_{t} p(E_t \mid B, \alpha_t),
\end{equation}
where $\alpha_t(x)$ denotes the spatial precision of task evidence $E_t$. This formulation corresponds to a Gaussian prior centered at $G$ with precision $w_0$ and Gaussian evidence likelihoods centered at $B$ with spatial precision $\alpha_t(x)$. \textcolor{blue}{The factorization is a conditional-independence working model for continuous latent evidence. It conditions on the realized reference and precisions during the local solve.} Ignoring constants and the factor $1/2$, the negative log-posterior is
\begin{equation}
\mathcal{L}(B) = w_0 \|B - G\|_2^2 + \sum_t \alpha_t \|B - E_t\|_2^2,
\end{equation}
where $w_0$ is the precision of the shared prior/reference. {\color{blue} It is predicted by $w_0=\mathrm{softplus}(\mathrm{Conv}_{1\times1}(G))+\epsilon$, with $\epsilon=10^{-6}$. The precisions $w_0$ and $\alpha_t$ are spatial scalars shared across feature channels. \par} The closed-form solution is
\begin{equation}
B = \frac{w_0 G + \sum_t \alpha_t E_t}{w_0 + \sum_t \alpha_t}.
\end{equation}
All multiplications and divisions in this expression are applied element-wise over spatial positions or tokens.

The closed-form bridge $B$ gives the posterior aggregation result before lightweight correction. As shown in Fig.~\ref{fig:pbo}(d), we further use a bounded posterior correction step
\begin{equation}
\widehat{B}=G+\eta_b(B-G), \quad 0<\eta_b<1,
\label{eq:pbo_correction}
\end{equation}
where $G$ is the shared reference and $\eta_b$ controls the bridge-correction strength.
Fig.~\ref{fig:pbo}(d) uses $\eta$ as a simplified notation for this bounded correction coefficient.
\textcolor{blue}{Each stage learns an independent sigmoid-parameterized $\eta_b$.
The corrected bridge $\widehat B$ lies on the segment from $G$ to $B$.}

{\color{blue}
\textbf{Proposition 1.}
For fixed evidence and positive precisions, let $D=w_0+\sum_t\alpha_t$.
The posterior bridge uniquely minimizes the local quadratic objective, and
\[
\begin{aligned}
\mathcal L(V)-\mathcal L(B)&=D\|V-B\|_2^2,\\
\mathcal L(\widehat B)-\mathcal L(B)
&=(1-\eta_b)^2[\mathcal L(G)-\mathcal L(B)].
\end{aligned}
\]
\emph{Proof.}
Expanding the objective and using $DB=w_0G+\sum_t\alpha_tE_t$ gives
\[
\mathcal L(V)=D\|V-B\|_2^2+\mathcal L(B).
\]
Since $D>0$, the minimum is attained uniquely at $V=B$.
Substituting $\widehat B-B=(1-\eta_b)(G-B)$ proves the correction identity.
\hfill$\square$
\par}

{\color{blue}
This deterministic identity holds for every realized evidence configuration with positive precisions.
The implemented denominator $D+\epsilon$ corresponds to minimizing
$\mathcal L_\epsilon(V)=\mathcal L(V)+\epsilon\|V\|_2^2$.
This term specifies the stabilized forward solver, while supervised optimization uses $\mathcal L_{\mathrm{train}}$.
Precision parameters remain differentiable during end-to-end training.
Relative to an ideal bridge $B^\star$, define $\varepsilon_0=G-B^\star$ and $\varepsilon_t=E_t-B^\star$.
For the exact normalized bridge, minimum-variance optimality among scalar-weighted
unbiased combinations requires conditionally zero-mean, mutually uncorrelated errors
with isotropic covariances $\sigma_j^2I$, $\sigma_j^2>0$, for $j=0,\ldots,T$.
The corresponding precisions satisfy $w_0=\kappa\sigma_0^{-2}$ and
$\alpha_t=\kappa\sigma_t^{-2}$ for one $\kappa>0$.
These moments are conditioned on information fixing the coefficients and ideal bridge.
Precision ordering supports relative weighting, while the variance-optimality result
requires quantitative inverse-variance calibration.
\par}

{\color{blue}
The effect of reference error is explicit in the exact normalized form.
Writing $\rho_0=w_0/D$ and $\rho_t=\alpha_t/D$ gives
\[
\widehat B=(1-\eta_b+\eta_b\rho_0)G+\eta_b\sum_t\rho_tE_t.
\]
With the coefficients held fixed, $1-\eta_b+\eta_b\rho_0$ determines the direct contribution
of reference information and reference error.
If $G=B^\star+b+\delta_0$ and $E_t=B^\star+b+\delta_t$, with shared bias $b$, then
\[
B-B^\star=b+\rho_0\delta_0+\sum_t\rho_t\delta_t.
\]
Thus, a common error component remains in normalized fusion, as can occur with a biased
shared representation or missing scene information across all branches.
The conditional error analysis is provided in the Supplementary Material.
\par}

{\color{blue}
After correction, a token-MLP residual and, for $k>1$, an aligned previous-stage
reference addition form the reference used for dispatch.
Spatial alignment and any required channel projection give $ref_t$ for each task.
This previous-stage addition occurs after PFE evaluates the current $G$.
In Sec.~\ref{sec:cdo}, $B$ denotes this actual aligned dispatch reference.
The segment property above applies to $\widehat B$ before subsequent refinement.
The parameter-free designation applies to the closed-form PBO aggregation core.
\par}

\subsection{Precision Field Estimation}

The posterior bridge depends on spatial precision. \textcolor{blue}{PFE maps local evidence descriptors to positive precision at each native evidence cell.}

We define
\begin{equation}
z_t(x)=\big[\mathrm{sim}(E_t(x),G(x)),\ \mathrm{tv}(E_t)(x)\big].
\end{equation}
Similarity measures alignment with the shared reference. \textcolor{blue}{We use channel-wise cosine similarity and sum the channel-averaged absolute horizontal and vertical neighbor differences for local variation. These cues provide a compact descriptor of reference agreement and spatial change.}

We use a rule-based mapping.
\begin{equation}
\mu_r(x) \textcolor{blue}{=} \exp\left(-\frac{1}{2}\left\|\frac{z_t(x)-c_r}{s_r}\right\|^2\right),
\end{equation}
and normalize the rule activations as
\begin{equation}
\bar{\mu}_r(x)=\frac{\mu_r(x)}{\sum_j \mu_j(x)+\epsilon}.
\end{equation}
\textcolor{blue}{Division by the positive scale vector $s_r$ is componentwise.
The symbols $\log\alpha_r$ and $\log\alpha_t$ denote the rule and aggregated
pre-Softplus scores in this mapping.}
For each rule, the log-precision is predicted by
\begin{equation}
\begin{aligned}
\log \alpha_r(x)
&=a_{\mathrm{sim}}^{(r)}\mathrm{sim}(E_t(x),G(x))\\
&\quad+a_{\mathrm{tv}}^{(r)}\mathrm{tv}(E_t)(x)+b_r.
\end{aligned}
\end{equation}
The final log-precision is aggregated as
\begin{equation}
\log \alpha_t(x)=\sum_r \bar{\mu}_r(x)\log \alpha_r(x),
\end{equation}
and the positive precision field is obtained by
\begin{equation}
\alpha_t(x)=\mathrm{softplus}(\log \alpha_t(x))\textcolor{blue}{+\epsilon}.
\end{equation}

{\color{blue}
Each stage uses eight rules with centers and scales in $\mathbb R^2$.
We initialize $c_{rj}\sim\mathcal N(0,0.2^2)$, the affine coefficients
$(a_{\mathrm{sim}}^{(r)},a_{\mathrm{tv}}^{(r)})=(3,-2)$, and $b_r=0$.
Scales are obtained by applying Softplus to zero-initialized raw parameters and adding
$\epsilon$, giving $s_{rj}^{init}=\log2+\epsilon$.
The three stage-level rule banks are learned independently, with parameters shared across tasks within each stage.
All rule parameters are optimized jointly with the network and fixed at inference,
while activations and precision remain input-dependent.
\par}

{\color{blue}
The default $1\times1$ precision block assigns one scalar to each restored native evidence cell,
shared across its channels.
For InternImage inputs of $448\times576$ on NYUD-v2, the three precision grids are
$14\times18$, $28\times36$ and $56\times72$.
For $512\times512$ PASCAL-Context inputs, they are
$16\times16$, $32\times32$ and $64\times64$.
Their image-space sampling strides are 32, 16 and 8, respectively.
The backbone and evidence extractor determine the contextual support of each cell.
\par}

\begin{figure}[t]
\centering
\subfloat[PFE.]{
    \includegraphics[width=0.85\columnwidth]{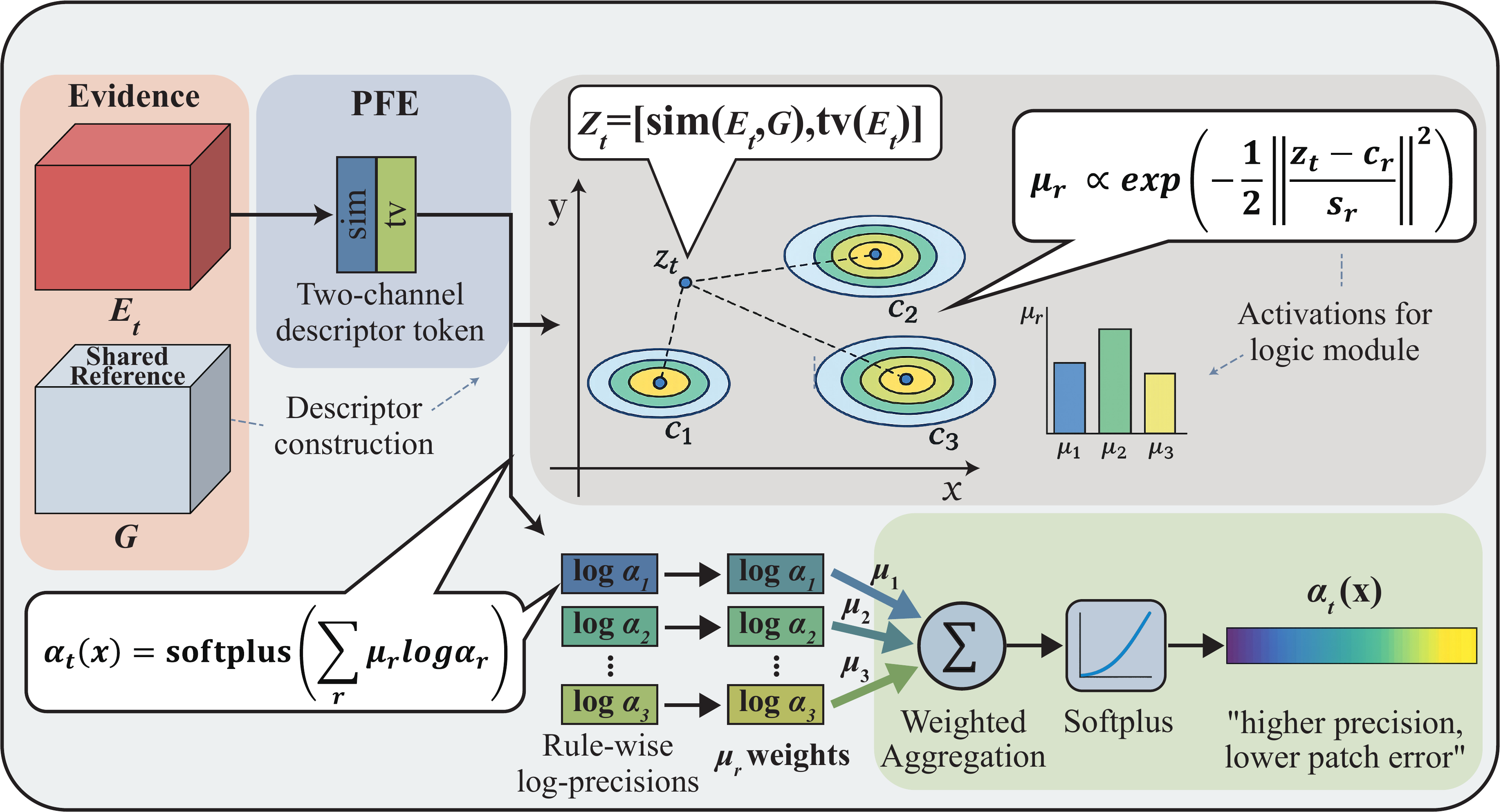}
    \label{fig:pfe_operator}
}
\\[-2pt]
\subfloat[CDO.]{
    \includegraphics[width=0.85\columnwidth]{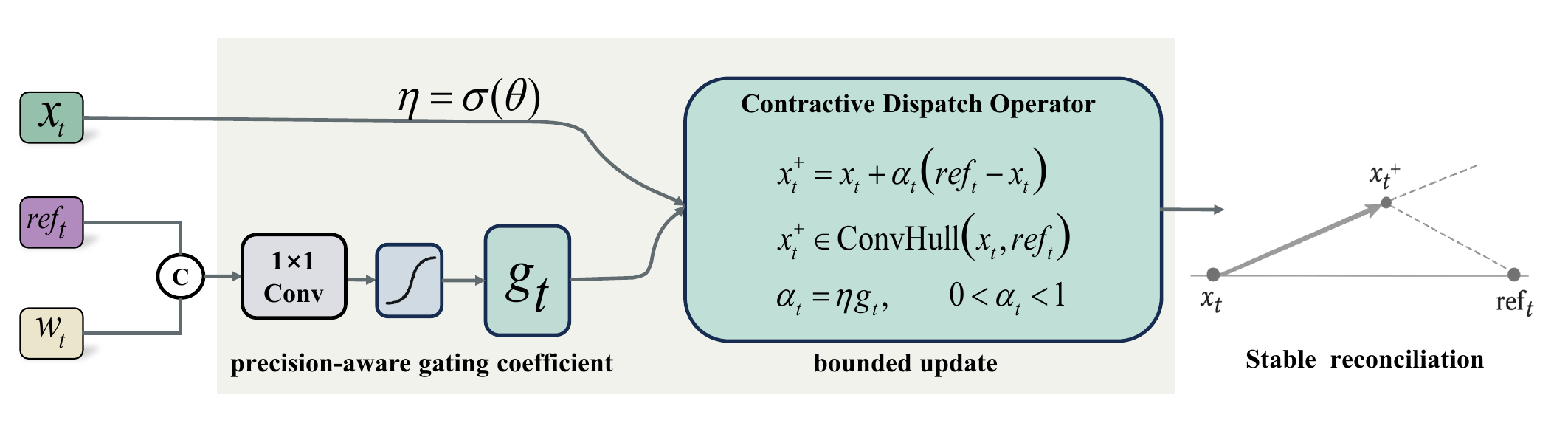}
    \label{fig:cdo_operator}
}
\vspace{-2pt}
\caption{\textbf{Illustration of PFE and CDO.}
(a) PFE estimates the spatial precision $\alpha_t(x)$ from task-reference similarity and local variation.
(b) CDO redistributes the bridge reference $ref_t$ to each task through the bounded update $x_t^{+}=x_t+\alpha_t(ref_t-x_t)$ shown in the diagram. Here, $\alpha_t$ denotes the effective update coefficient in the conceptual illustration. In the text, this coefficient is written as $\beta_t=\eta g_t$ to distinguish it from the PFE precision field used for posterior bridge construction. \textcolor{blue}{The geometric illustration applies componentwise to the implemented channel-wise gate.}}
\label{fig:pfe_cdo}
\end{figure}

\subsection{Contractive Dispatch Operator}
\label{sec:cdo}

Given a posterior bridge $B$, direct overwrite may suppress task-specific structures and propagate unreliable shared information. We therefore redistribute the bridge through a bounded bridge-to-task update. Here, $B$ denotes the aligned bridge reference used for dispatch. This corresponds to $ref_t$ in Fig.~\ref{fig:pfe_cdo}(b), where the subscript emphasizes that the bridge reference is dispatched to a specific task branch.

For each task, CDO takes the current task state $X_t$, the bridge reference $B$, and the precision field $\alpha_t$ estimated by PFE as inputs. In Fig.~\ref{fig:pfe_cdo}(b), $w_t$ denotes the precision-related dispatch cue derived from PFE. The figure uses $\alpha_t$ to denote the effective local update coefficient in the simplified form $x_t^{+}=x_t+\alpha_t(ref_t-x_t)$. In the following equations, $\beta_t(x)$ denotes the effective dispatch coefficient and $\alpha_t(x)$ denotes PFE precision. The dispatch gate is computed as
\begin{equation}
\textcolor{blue}{g_t(x)=\sigma\left(W_t*[B(x),\alpha_t(x)]+b_t^g\right),}
\end{equation}
where $\sigma(\cdot)$ denotes the sigmoid function.
\textcolor{blue}{For $C$ task-feature channels, $W_t$ is a $1\times1$ convolution with $C+1$ inputs
and $C$ outputs, and $b_t^g$ is its bias.
A task-stage-specific step parameter is defined as}
\begin{equation}
\eta=\sigma(\theta).
\end{equation}
\textcolor{blue}{The parameters $W_t,b_t^g,\theta$ are learned separately for each task-stage instance
and reused across its refinement calls.
The stage index of $W_t,b_t^g$ and task-stage indices of $\eta,\theta$ are suppressed here.}
The effective local update coefficient is
\begin{equation}
\beta_t(x)=\eta g_t(x), \quad \textcolor{blue}{0<\beta_{t,c}(x)<\eta<1.}
\end{equation}
The task state is updated by
\begin{equation}
X_t^{+}(x)=X_t(x)+\beta_t(x)\textcolor{blue}{\odot}\left(B(x)-X_t(x)\right).
\end{equation}
\textcolor{blue}{Here $\odot$ denotes element-wise multiplication.}

{\color{blue}
\textbf{Proposition 2.}
For one dispatch with a fixed reference and $0<\beta_{t,c}(x)<1$,
each updated component lies between its current value and the corresponding reference value.
\par}
At each spatial location, the update satisfies
\begin{equation}
X_t^{+}(x)-B(x)=\left(1-\beta_t(x)\right)\textcolor{blue}{\odot}\left(X_t(x)-B(x)\right).
\end{equation}
\textcolor{blue}{Consequently, the bridge-relative deviation satisfies}
\begin{equation}
\|X_t^{+}(x)-B(x)\|_2 \leq \|X_t(x)-B(x)\|_2.
\end{equation}

{\color{blue}
\emph{Proof.}
Suppressing $x$, each component is
\[
X_{t,c}^+=(1-\beta_{t,c})X_{t,c}+\beta_{t,c}B_c.
\]
The positive coefficients sum to one, proving the componentwise interval property.
Moreover,
\[
\begin{aligned}
\|X_t^+-B\|_2^2
&=\sum_c(1-\beta_{t,c})^2(X_{t,c}-B_c)^2\\
&\leq\|X_t-B\|_2^2.
\end{aligned}
\]
Taking square roots establishes the bound.
\hfill$\square$
\par}

{\color{blue}
The vector lies on the endpoint segment when all channels share one coefficient.
For channel-wise gates, the convex-hull illustration in Fig.~\ref{fig:pfe_cdo}(b)
applies to each component.
The guarantee concerns one dispatch before reference concatenation and nonlinear refinement.
Global convergence additionally requires assumptions on the recomputed references,
input-dependent gates and subsequent transformations.
\par}

We embed this update into a multi-stage decoder.
\begin{equation}
\textcolor{blue}{
\begin{aligned}
X_t^+&=\mathrm{CDO}(X_t^{(k-1)},ref_t^{(k)},\alpha_t^{(k)}),\\
X_t^{(k)}&=\mathrm{Refine}_{t,k}(X_t^{(k-1)},X_t^+,ref_t^{(k)}),
\end{aligned}}
\end{equation}
where
\begin{equation}
B^{(k)} = \mathrm{PBO}(G^{(k)}, \{E_t^{(k)}\}, \{\alpha_t^{(k)}\}).
\end{equation}
\textcolor{blue}{For $k=1,2,3$, the corrected and refined $B^{(k)}$ is aligned to obtain $ref_t^{(k)}$.
The map $\mathrm{Refine}_{t,k}$ concatenates the CDO output with the reference and applies
convolutional refinement with a residual connection to the current task state.
The two task-update operations repeat within each stage using the same CDO parameters.}
The stage outputs are aggregated by
\begin{equation}
\widetilde{X}_t = \sum_{k=1}^{3} \phi_k(X_t^{(k)}), \quad
\hat{Y}_t = h_t(\widetilde{X}_t).
\end{equation}

This design performs progressive posterior bridge propagation. \textcolor{blue}{The stages operate at progressively finer resolutions. PFE controls local evidence weighting, PBO constructs the shared posterior, and CDO bounds each dispatch before task refinement. Together, these operations organize evidence contribution and task adaptation within the decoder.}

\section{Experiments}
\subsection{Experimental Setup}
\textbf{Datasets.}
We evaluate $\mathcal{B}^{3}$-Net on three standard multi-task dense prediction benchmarks, including NYUD-v2~\cite{NYUDv2}, PASCAL-Context~\cite{PASCALContext}, and Cityscapes~\cite{Cityscapes}. Following the commonly used protocols in dense multi-task prediction~\cite{InvPT,InvPTPlusPlus,BridgeNet,MTMamba,MTMambaPlusPlus}, NYUD-v2 is evaluated on four tasks, including semantic segmentation, depth estimation, surface normal estimation, and edge detection. PASCAL-Context is evaluated on five tasks, including semantic segmentation, human parsing, saliency estimation, surface normal estimation, and edge detection. Cityscapes is evaluated on semantic segmentation and depth estimation. We keep the data splits, preprocessing, task definitions, label usage, and evaluation protocol consistent with prior works whenever applicable~\cite{InvPT,InvPTPlusPlus,BridgeNet,MTMamba,MTMambaPlusPlus}.

\newcommand{\best}[1]{\textbf{#1}}
\newcommand{\second}[1]{\underline{#1}}
\begin{table*}[t]
\centering
\small
\setlength{\tabcolsep}{6.5pt}
\renewcommand{\arraystretch}{1.12}
\caption{Comparison with representative state-of-the-art methods under their reported settings on NYUD-v2 (left)  and PASCAL-Context (right). $\uparrow$ ($\downarrow$) indicates that a higher (lower) result corresponds to better performance. The best and second best results are highlighted in \best{bold} and \second{underline}, respectively.}
\label{tab:sota_nyud_pascal_bridge_feature_decoder}
\vspace{2pt}

\resizebox{\textwidth}{!}{%
\begin{tabular}{lcccc@{\hspace{14pt}}lccccc}
\toprule
\textbf{Method} &
\textbf{Semseg} & \textbf{Depth} & \textbf{Normal} & \textbf{Edge}
& \textbf{Method} &
\textbf{Semseg} & \textbf{Parsing} & \textbf{Saliency} & \textbf{Normal} & \textbf{Edge} \\
& mIoU$\uparrow$ & RMSE$\downarrow$ & mErr$\downarrow$ & odsF$\uparrow$
& & mIoU$\uparrow$ & mIoU$\uparrow$ & maxF$\uparrow$ & mErr$\downarrow$ & odsF$\uparrow$ \\
\cmidrule(lr){1-5}\cmidrule(lr){6-11}

\multicolumn{5}{c}{\textit{CNN-based decoder}} &
\multicolumn{6}{c}{\textit{CNN-based decoder}} \\
\cmidrule(lr){1-5}\cmidrule(lr){6-11}
Cross-Stitch~\cite{CrossStitch} & 36.34 & 0.6290 & 20.88 & 76.38 &
ASTMT~\cite{ASTMT}        & 68.00 & 61.10 & 65.70 & 14.70 & 72.40 \\
PAP~\cite{PAP}          & 36.72 & 0.6178 & 20.82 & 76.42 &
PAD-Net~\cite{PADNet}      & 53.60 & 59.60 & 65.80 & 15.30 & 72.50 \\
PSD~\cite{PSD}          & 36.69 & 0.6246 & 20.87 & 76.42 &
MTI-Net~\cite{MTINet}      & 61.70 & 60.18 & 84.78 & 14.23 & 70.80 \\
PAD-Net~\cite{PADNet}      & 36.61 & 0.6270 & 20.85 & 76.38 &
ATRC~\cite{ATRC}         & 62.69 & 59.42 & 84.70 & 14.20 & 70.96 \\
MTI-Net~\cite{MTINet}      & 45.97 & 0.5365 & 20.27 & 77.86 &
ATRC-ASPP~\cite{ATRC}    & 63.60 & 60.23 & 83.91 & 14.30 & 70.86 \\
ATRC~\cite{ATRC}         & 46.33 & 0.5363 & 20.18 & 77.94 &
ATRC-BMTAS~\cite{ATRC}   & 67.67 & 62.93 & 82.29 & 14.24 & 72.42 \\

\cmidrule(lr){1-5}\cmidrule(lr){6-11}
\multicolumn{5}{c}{\textit{Transformer-based decoder}} &
\multicolumn{6}{c}{\textit{Transformer-based decoder}} \\
\cmidrule(lr){1-5}\cmidrule(lr){6-11}
InvPT~\cite{InvPT}           & 53.56 & 0.5183 & 19.04 & 78.10 &
InvPT~\cite{InvPT}           & 79.03 & 67.61 & 84.81 & 14.15 & 73.00 \\
InvPT++~\cite{InvPTPlusPlus}         & 53.85 & 0.5096 & 18.67 & 78.10 &
InvPT++~\cite{InvPTPlusPlus}         & 80.22 & 69.12 & 84.74 & 13.73 & 74.20 \\
TaskPrompter~\cite{TaskPrompter}    & 55.30 & 0.5152 & 18.47 & 78.20 &
TaskPrompter~\cite{TaskPrompter}    & 80.89 & 68.89 & 84.83 & 13.72 & 73.50 \\
MQTransformer~\cite{MQTransformer}   & 54.84 & 0.5325 & 19.67 & 78.20 &
MQTransformer~\cite{MQTransformer}   & 78.93 & 67.41 & 83.58 & 14.21 & 73.90 \\
TSP-Transformer~\cite{TSPTransformer} & 55.39 & 0.4961 & 18.44 & 77.50 &
TSP-Transformer~\cite{TSPTransformer} & \second{81.48} & 70.64 & 84.86 & 13.69 & 74.80 \\
MLoRE~\cite{MLoRE}           & 55.96 & 0.5076 & 18.33 & 78.43 &
MLoRE~\cite{MLoRE}           & 81.41 & 70.52 & 84.90 & 13.51 & 75.42 \\

\cmidrule(lr){1-5}\cmidrule(lr){6-11}
\multicolumn{5}{c}{\textit{Diffusion-based decoder}} &
\multicolumn{6}{c}{\textit{Diffusion-based decoder}} \\
\cmidrule(lr){1-5}\cmidrule(lr){6-11}
TaskDiffusion~\cite{TaskDiffusion}   & 55.65 & 0.5020 & 18.43 & 78.64 &
TaskDiffusion~\cite{TaskDiffusion}   & 81.21 & 69.62 & 84.94 & 13.55 & 74.89 \\

\cmidrule(lr){1-5}\cmidrule(lr){6-11}
\multicolumn{5}{c}{\textit{Mamba-based decoder}} &
\multicolumn{6}{c}{\textit{Mamba-based decoder}} \\
\cmidrule(lr){1-5}\cmidrule(lr){6-11}
MTMamba~\cite{MTMamba}         & 55.82 & 0.5066 & 18.63 & 78.70 &
MTMamba~\cite{MTMamba}         & 81.11 & 72.62 & 84.14 & 14.14 & \second{78.80} \\
MTMamba++~\cite{MTMambaPlusPlus}       & \second{57.01} & 0.4818 & 18.27 & 79.40 &
MTMamba++~\cite{MTMambaPlusPlus}       & \best{81.94} & \second{72.87} & 85.56 & 14.29 & 78.60 \\

\cmidrule(lr){1-5}\cmidrule(lr){6-11}
\multicolumn{5}{c}{\textit{Bridge-feature-based decoder}} &
\multicolumn{6}{c}{\textit{Bridge-feature-based decoder}} \\
\cmidrule(lr){1-5}\cmidrule(lr){6-11}
BridgeNet~\cite{BridgeNet}         & 56.57 & \second{0.4655} & \second{17.29} & \second{80.02} &
BridgeNet~\cite{BridgeNet}         & 79.89 & 71.33 & \second{85.64} & \best{13.38} & 73.24 \\
\rowcolor{oursblue}
B$^{3}$-Net (Ours) & \best{57.78} & \best{0.4587} & \best{17.22} & \best{83.18} &
B$^{3}$-Net (Ours) & 80.81 & \best{73.73} & \best{86.11} & \second{13.40} & \best{81.18} \\
\bottomrule
\end{tabular}%
}
\end{table*}

\textbf{Evaluation metrics.}
We adopt the standard task-specific metrics used in previous dense multi-task prediction methods~\cite{InvPT,InvPTPlusPlus,BridgeNet,MTMamba,MTMambaPlusPlus}. Semantic segmentation and human parsing are evaluated by mean intersection over union. Depth estimation is evaluated by RMSE on NYUD-v2 and by Abs. Rel. and Sq. Rel. errors on Cityscapes. Surface normal estimation is evaluated by mean angular error. Saliency estimation is evaluated by maximal F-measure. Edge detection is evaluated by optimal dataset scale F-measure. To measure the overall multi-task trade-off, we report the task-wise transfer gain $\Delta_{\tau}(\%)$ and the averaged multi-task transfer gain $\Delta_{\mathrm{MTL}}(\%)$ against single-task baselines:
\begin{equation}
\Delta_{\tau}(\%)=
\begin{cases}
\frac{M^{MT}_{\tau}-M^{ST}_{\tau}}{M^{ST}_{\tau}}\times100, & \tau:\uparrow \\
\frac{M^{ST}_{\tau}-M^{MT}_{\tau}}{M^{ST}_{\tau}}\times100, & \tau:\downarrow
\end{cases}
\label{eq:deltatau}
\end{equation}
\begin{equation}
\Delta_{\mathrm{MTL}}(\%)=\frac{1}{|\mathcal{T}|}\sum_{\tau\in\mathcal{T}}\Delta_{\tau}(\%).
\label{eq:deltamtl}
\end{equation}
Here, $M^{MT}_{\tau}$ and $M^{ST}_{\tau}$ denote the performance of the multi-task model and the corresponding single-task baseline on task $\tau$, respectively. The notation $\tau:\uparrow$ denotes metrics where larger values are better, while $\tau:\downarrow$ denotes metrics where smaller values are better.

\textbf{Implementation details.}
All experiments are implemented in PyTorch and conducted on NVIDIA A100 40G GPUs. For fair comparison, the default backbone selection follows~\cite{BridgeNet}. Specifically, InternImage-Large is used as the default backbone on NYUD-v2 and PASCAL-Context, and ViT-Large is used as the default backbone on Cityscapes.  We further keep the data processing, task definitions, evaluation metrics, and comparison protocol aligned with representative dense multi-task methods, including InvPT, InvPT++, MTMamba, and MTMamba++~\cite{InvPT,InvPTPlusPlus,MTMamba,MTMambaPlusPlus}. The recommended environment is built with Python 3.8, PyTorch 1.13.1 with CUDA 11.7, torchvision 0.14.1, timm 0.5.4, einops 0.4.1, mmsegmentation 0.30.0, mmcv 1.7.0, numpy 1.26.4, and DCNv3 operators compiled for InternImage backbones.

\textbf{Training protocol.}
Following the training protocol for InternImage backbones whenever applicable~\cite{BridgeNet}, we use AdamW as the optimizer and adopt a polynomial learning rate scheduler.
AdamW uses $(\beta_1,\beta_2)=(0.9,0.999)$.
The polynomial schedule has power $0.9$ and decays to zero over the training budget, measured in optimizer updates.
For NYUD-v2, the model is trained with a batch size of 3.
The learning rate is set to $5.0\times10^{-5}$, the weight decay is set to 0.03, and the training budget is 26K updates.
For PASCAL-Context, the model is trained with a batch size of 4.
The learning rate is set to $3.0\times10^{-5}$, the weight decay is set to 0.03, and the training budget is 45K updates.
For Cityscapes, the model is trained with a batch size of 3.
The learning rate is set to $4.0\times10^{-5}$, the weight decay is set to 0.008, and the training budget is 40K updates.
Gradient clipping is used for stable training, with maximum norms of 10, 5, and 7 for NYUD-v2, PASCAL-Context, and Cityscapes, respectively.
All configurations use a validation batch size of 6 and are evaluated every 1K updates and at the final update.
Backbone-matched comparisons with ViT-B, ViT-L, and InternImage-B use their corresponding backbone-specific configuration files (see code). The accompanying code also provides shorter InternImage-L schedules
of 20K updates on NYUD-v2 and 40K on PASCAL-Context, both with
batch size 4, to accommodate different computational budgets.
On NYUD-v2, the larger batch keeps the number of training image
presentations comparable to the default schedule, with 80K
versus 78K.
On PASCAL-Context, the shorter schedule reduces image presentations
from 180K to 160K at the same batch size.

\textbf{Baselines and comparison protocol.}
We compare $\mathcal{B}^{3}$-Net with representative CNN-based, Transformer-based, diffusion-based, Mamba-based, and bridge-feature-based dense multi-task methods, including Cross-Stitch, PAP, PSD, PAD-Net, MTI-Net, ATRC, InvPT, InvPT++, TaskPrompter, MQTransformer, TSP-Transformer, MLoRE, TaskDiffusion, MTMamba, MTMamba++, and BridgeNet~\cite{CrossStitch,PAP,PSD,PADNet,MTINet,ATRC,InvPT,InvPTPlusPlus,TaskPrompter,MQTransformer,TSPTransformer,MLoRE,TaskDiffusion,MTMamba,MTMambaPlusPlus,BridgeNet}. Reported-setting comparisons follow the evaluation settings of prior works. Backbone-matched comparisons are further conducted with ViT-Base, ViT-Large, InternImage-Base, and InternImage-Large to separate the contribution of the proposed decoder interaction mechanism from backbone capacity.

\subsection{Comparison with State-of-the-Art Methods}
\textbf{Reported-setting comparison.}
Table~\ref{tab:sota_nyud_pascal_bridge_feature_decoder} reports the comparison with representative dense multi-task methods on NYUD-v2 and PASCAL-Context under their reported settings. The compared methods cover CNN-based, Transformer-based, diffusion-based, Mamba-based, and bridge-feature-based decoders. On NYUD-v2, $\mathcal{B}^{3}$-Net achieves the best results on all four tasks. It obtains 57.78 mIoU for semantic segmentation, 0.4587 RMSE for depth estimation, 17.22 mErr for surface normal estimation, and 83.18 odsF for edge detection. Compared with MTMamba++~\cite{MTMambaPlusPlus}, $\mathcal{B}^{3}$-Net improves semantic segmentation by 0.77 mIoU, reduces depth RMSE by 0.0231, reduces normal mErr by 1.05, and improves edge odsF by 3.78. Compared with BridgeNet~\cite{BridgeNet}, $\mathcal{B}^{3}$-Net also improves all four metrics. This shows that the proposed controlled posterior bridge is more effective than deterministic bridge-feature interaction for balancing semantic, geometric, and boundary tasks.

On PASCAL-Context, $\mathcal{B}^{3}$-Net achieves the best results on human parsing, saliency estimation, and edge detection, with 73.73 mIoU, 86.11 maxF, and 81.18 odsF, respectively. It also gives competitive performance on semantic segmentation and surface normal estimation. This dataset contains five heterogeneous tasks and usually induces stronger task competition than NYUD-v2. The gains on parsing, saliency, and edge detection indicate that $\mathcal{B}^{3}$-Net improves the transfer of foreground, part-level, and boundary cues while maintaining a balanced multi-task trade-off.

\begin{table}[t]
\centering
\small
\setlength{\tabcolsep}{8pt}
\renewcommand{\arraystretch}{1.10}
\caption{\textbf{Quantitative comparison on Cityscapes.}
Segmentation is reported as mIoU in percentage form without the percent symbol.
Depth is evaluated by Abs. Rel.~($\downarrow$) and Sq. Rel.~($\downarrow$).
The best results are highlighted in bold.}
\label{tab:cityscapes_comparison}
\begin{tabular}{lccc}
\toprule
\multicolumn{4}{c}{Cityscapes} \\
\midrule
\textbf{Method} & \textbf{Semseg} & \multicolumn{2}{c}{\textbf{Depth}} \\
\cmidrule(lr){2-2} \cmidrule(lr){3-4}
& mIoU$\uparrow$ & Abs. Rel.$\downarrow$ & Sq. Rel.$\downarrow$ \\
\midrule
MTAN~\cite{MTAN}                    & 88.90            & 0.1435            & 1.622            \\
PAD-Net~\cite{PADNet}                 & 85.59            & 0.1423            & 1.671            \\
InvPT~\cite{InvPT}                   & 89.86            & 0.1808            & 1.907            \\
SwinMTL~\cite{SwinMTL}                 & 85.04            & 0.1518            & 1.941            \\
TaskPrompter~\cite{TaskPrompter}            & 85.26            & 0.1962            & 2.689            \\
MTMamba~\cite{MTMamba}                 & 90.77            & 0.1169            & 1.355            \\
MTMamba++~\cite{MTMambaPlusPlus}               & 91.11            & 0.1131            & 1.239            \\
BridgeNet~\cite{BridgeNet}               & 93.72            & 0.0899            & 0.850            \\
\rowcolor{oursblue}
B$^{3}$-Net (\textit{ours}) & \best{93.95} & \best{0.0889} & \best{0.842} \\
\bottomrule
\end{tabular}
\end{table}

Table~\ref{tab:cityscapes_comparison} further reports the results on Cityscapes. $\mathcal{B}^{3}$-Net achieves 93.95 mIoU for semantic segmentation, 0.0889 Abs. Rel. error, and 0.842 Sq. Rel. error for depth estimation. It outperforms MTAN~\cite{MTAN}, PAD-Net~\cite{PADNet}, InvPT~\cite{InvPT}, SwinMTL~\cite{SwinMTL}, TaskPrompter~\cite{TaskPrompter}, MTMamba~\cite{MTMamba}, MTMamba++~\cite{MTMambaPlusPlus}, and BridgeNet~\cite{BridgeNet}. Compared with BridgeNet~\cite{BridgeNet}, $\mathcal{B}^{3}$-Net improves segmentation from 93.72 to 93.95 and reduces Abs. Rel. and Sq. Rel. errors from 0.0899 and 0.850 to 0.0889 and 0.842. The gains in semantic and depth prediction extend the evidence for posterior bridge learning to urban scene understanding.

\textbf{Backbone-matched comparison.}
Table~\ref{tab:vit_and_intern_compare} compares decoder performance with matched backbones. On ViT-Base, $\mathcal{B}^{3}$-Net consistently outperforms InvPT~\cite{InvPT} and BridgeNet~\cite{BridgeNet} on NYUD-v2, achieving 52.82 mIoU, 0.5117 RMSE, 18.66 mErr, and 79.24 odsF. On PASCAL-Context, it also gives the best results on semantic segmentation, human parsing, saliency estimation, and edge detection within the ViT-Base group.

\begin{table*}[t]
\centering
\scriptsize
\setlength{\tabcolsep}{3.8pt}
\renewcommand{\arraystretch}{1.05}
\caption{\textbf{Comparison with backbone-matched multi-task baselines on NYUD-v2 and PASCAL-Context.}
Results with InternImage backbones are additionally reported at the bottom as stronger-backbone extensions.}
\label{tab:vit_and_intern_compare}
\resizebox{\textwidth}{!}{%
\begin{tabular}{ll cccc ccccc}
\toprule
& & \multicolumn{4}{c}{\textbf{NYUD-v2}} & \multicolumn{5}{c}{\textbf{PASCAL-Context}} \\
\cmidrule(lr){3-6}\cmidrule(lr){7-11}
\textbf{Backbone} & \textbf{Method}
& \textbf{Semseg} & \textbf{Depth} & \textbf{Normal} & \textbf{Edge}
& \textbf{Semseg} & \textbf{Parsing} & \textbf{Saliency} & \textbf{Normal} & \textbf{Edge} \\
&
& mIoU$\uparrow$ & RMSE$\downarrow$ & mErr$\downarrow$ & odsF$\uparrow$
& mIoU$\uparrow$ & mIoU$\uparrow$ & maxF$\uparrow$ & mErr$\downarrow$ & odsF$\uparrow$ \\
\midrule

\multirow{3}{*}{ViT-Base}
& InvPT~\cite{InvPT}
& 50.30 & 0.5367 & 19.00 & 77.86
& 77.33 & 66.62 & \underline{85.14} & 13.78 & \underline{73.20} \\
& BridgeNet~\cite{BridgeNet}
& \underline{51.14} & \underline{0.5186} & \underline{18.92} & \underline{77.98}
& \underline{77.98} & \underline{68.19} & 85.06 & \textbf{13.48} & 72.98 \\
& Ours
& \textbf{52.82} & \textbf{0.5117} & \textbf{18.66} & \textbf{79.24}
& \textbf{78.92} & \textbf{68.86} & \textbf{85.15} & \underline{13.70} &  \textbf{81.68} \\
\midrule

\multirow{4}{*}{ViT-Large}
& InvPT~\cite{InvPT}
& 53.56 & 0.5183 & 19.04 & 78.10
& 79.03 & 67.61 & \underline{84.81} & 14.15 & 73.00 \\
& TaskPrompter~\cite{TaskPrompter}
& \underline{55.30} & 0.5152 & \underline{18.47} & 78.20
& \underline{80.89} & 68.89 & \textbf{84.83} & \underline{13.72} & \underline{73.50} \\
& BridgeNet~\cite{BridgeNet}
& \textbf{55.51} & \textbf{0.4930} & \underline{18.47} & \underline{78.22}
& 80.64 & \underline{70.06} & 84.64 & \textbf{13.48} & 72.98 \\
& Ours
& 55.46 & \underline{0.4949} & \textbf{18.38} & \textbf{79.75}
& \textbf{81.03} & \textbf{71.33} & 84.57 & 13.85 & \textbf{80.92} \\
\midrule
\multicolumn{11}{c}{\textit{Stronger-backbone extension with InternImage}} \\
\midrule

\multirow{2}{*}{InternImage-B}
& BridgeNet~\cite{BridgeNet}
& -- & -- & -- & --
& 78.07 & 69.03 & \underline{85.29} & \textbf{13.44} & \underline{73.10} \\
& Ours
& \textbf{53.77} & \textbf{0.5000} & \textbf{18.92} & \textbf{82.01}
& \textbf{78.97} & \textbf{70.19} & \textbf{85.31} & \underline{13.78} & \textbf{80.32} \\
\midrule

\multirow{2}{*}{InternImage-L}
& BridgeNet~\cite{BridgeNet}
& -- & -- & -- & --
& \underline{79.89} & \underline{71.33} & \underline{85.64} & \textbf{13.38} & \underline{73.24} \\
& Ours
& \textbf{57.78} & \textbf{0.4587} & \textbf{17.22} & \textbf{83.18}
& \textbf{80.81} & \textbf{73.73} & \textbf{86.11} & \underline{13.40} & \textbf{81.18} \\
\bottomrule
\end{tabular}}
\end{table*}

\begin{figure}[t]
\centering
\setlength{\qualheight}{0.33\columnwidth}
\captionsetup[subfloat]{
    font=footnotesize,
    labelfont={rm,bf},
    textfont={rm,bf},
    justification=centering,
    singlelinecheck=true
}
\subfloat[NYUD-v2\label{fig:qualitative_nyud}]{%
    \begin{minipage}[b]{0.49\columnwidth}
        \centering
        \includegraphics[
            height=\qualheight,
            keepaspectratio
        ]{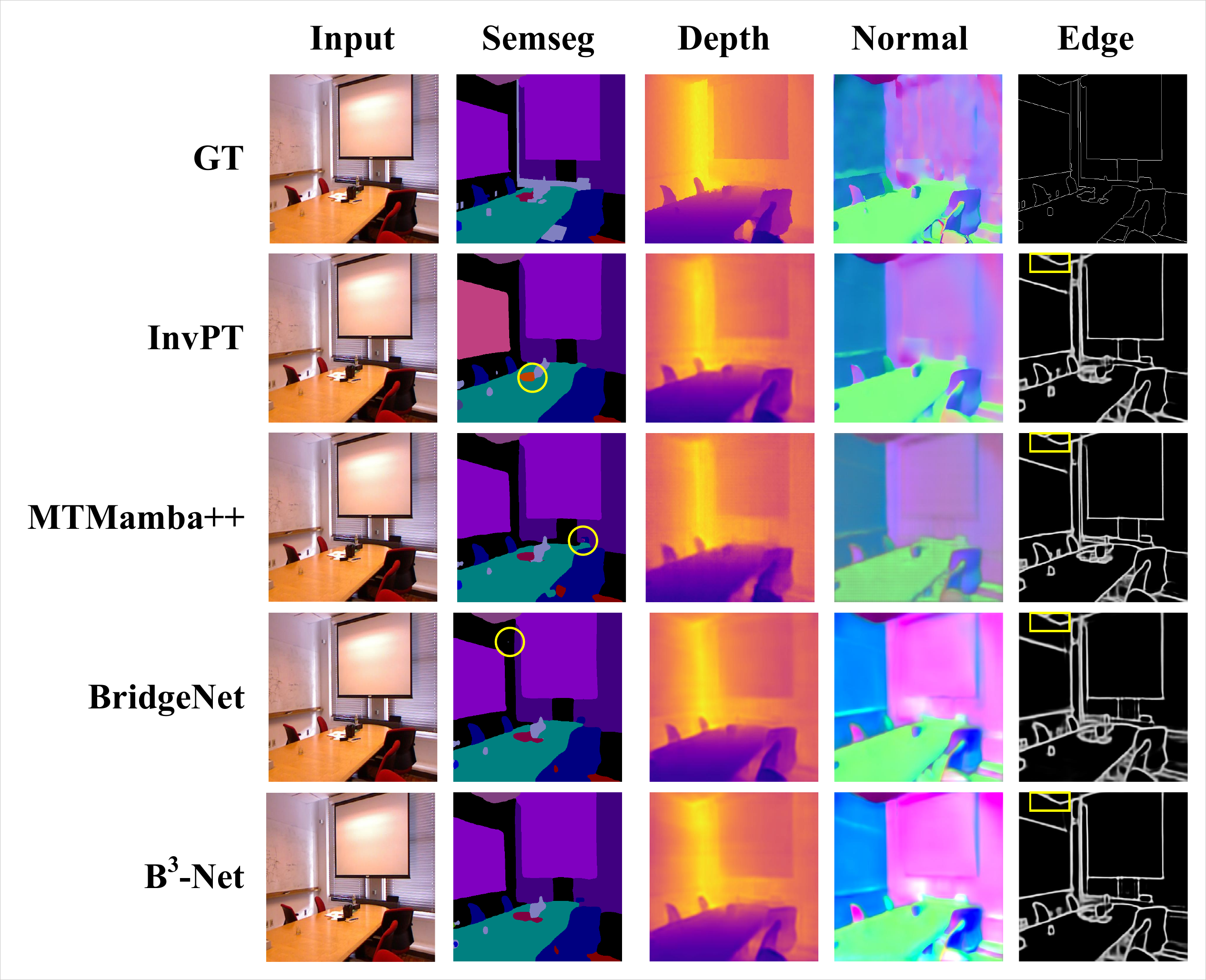}
    \end{minipage}%
}\hfill
\subfloat[PASCAL-Context\label{fig:qualitative_pascal}]{%
    \begin{minipage}[b]{0.49\columnwidth}
        \centering
        \includegraphics[
            height=\qualheight,
            keepaspectratio
        ]{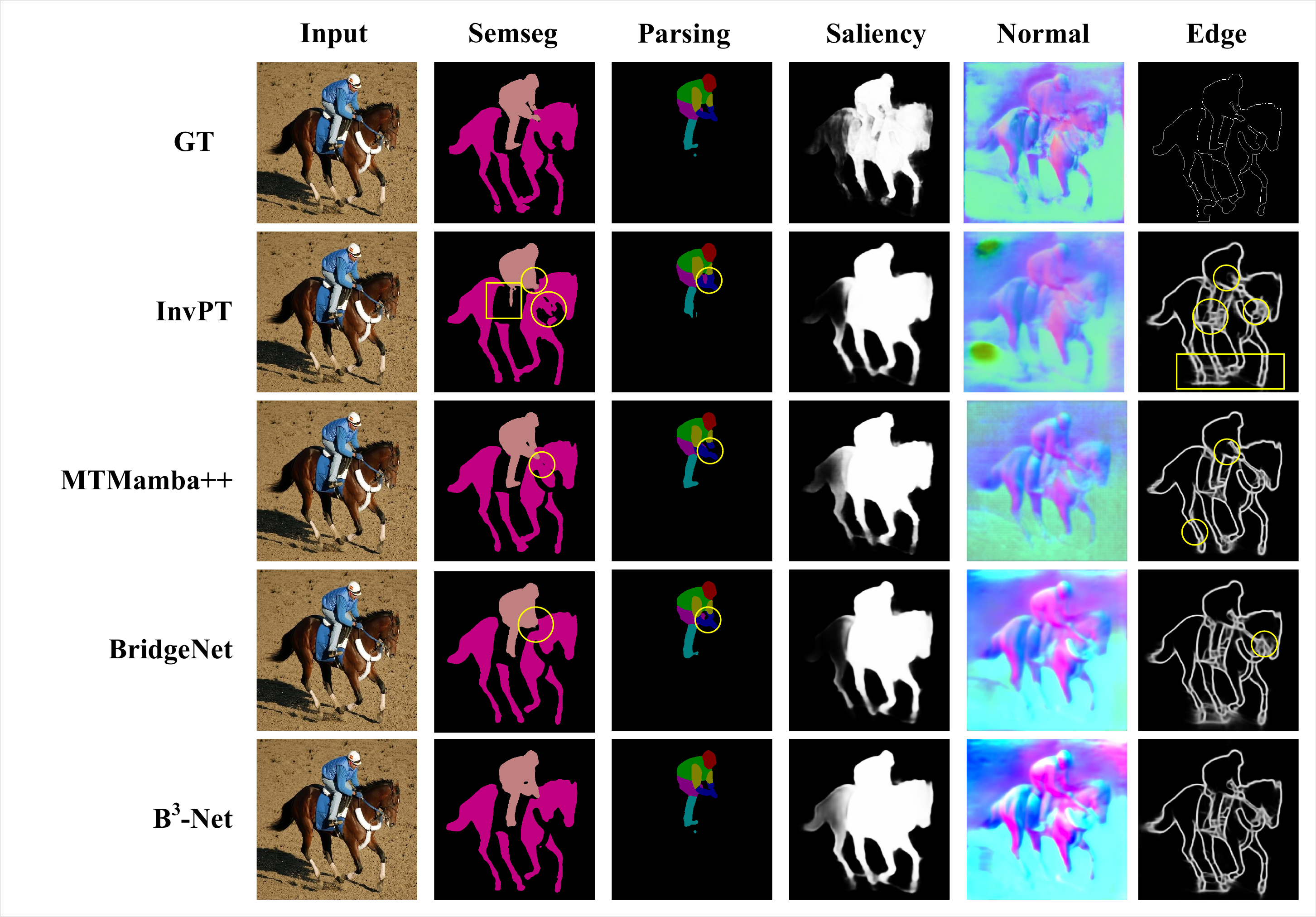}
    \end{minipage}%
}
\par\smallskip
\subfloat[Task-output heatmaps\label{fig:output_visualization}]{%
    \includegraphics[width=0.75\columnwidth]{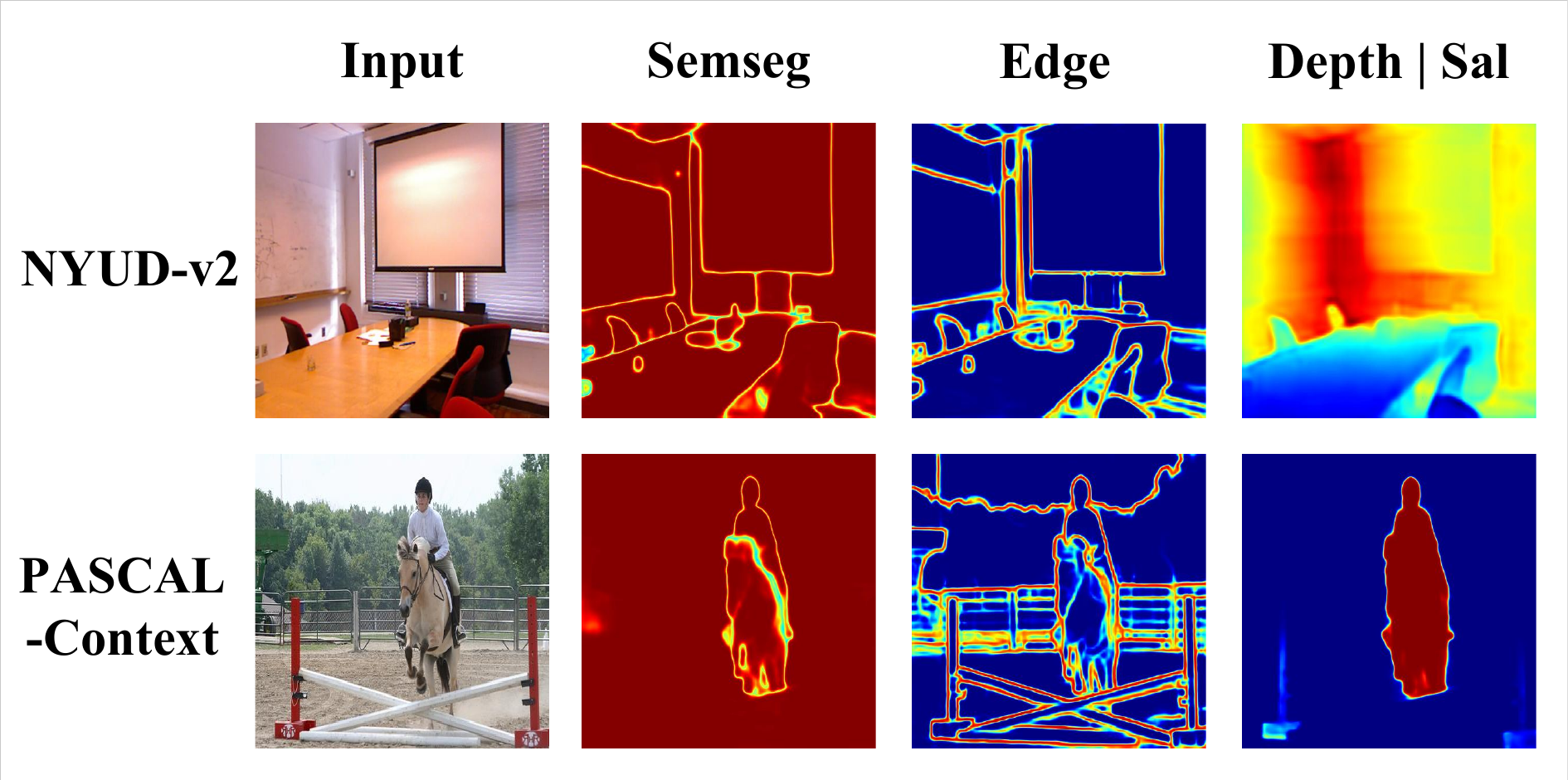}%
}
\caption{\textbf{Qualitative comparison and task-output heatmaps.}
(a,b) Comparisons with InvPT~\cite{InvPT},
MTMamba++~\cite{MTMambaPlusPlus}, and BridgeNet~\cite{BridgeNet}
on NYUD-v2 and PASCAL-Context, respectively.
Highlighted regions indicate representative errors of competing methods.
(c) Task-output heatmaps of B$^{3}$-Net on the NYUD-v2
and PASCAL-Context datasets.}
\label{fig:qualitative_comparison}
\end{figure}

On ViT-Large, $\mathcal{B}^{3}$-Net remains competitive against InvPT~\cite{InvPT}, TaskPrompter~\cite{TaskPrompter}, and BridgeNet~\cite{BridgeNet}. It obtains the best normal and edge performance on NYUD-v2, and the best semantic segmentation, human parsing, and edge detection performance on PASCAL-Context. The improvement on edge detection is particularly clear. $\mathcal{B}^{3}$-Net reaches 80.92 odsF, while TaskPrompter~\cite{TaskPrompter} and BridgeNet~\cite{BridgeNet} obtain 73.50 and 72.98, respectively. This suggests that the bounded bridge-to-task redistribution helps preserve local discontinuities and boundary structures.

The InternImage results further validate the scalability of the proposed decoder. With InternImage-Base and InternImage-Large, $\mathcal{B}^{3}$-Net achieves consistent gains over BridgeNet~\cite{BridgeNet} on most tasks. In particular, with InternImage-Large, our method obtains 57.78 mIoU, 0.4587 RMSE, 17.22 mErr, and 83.18 odsF on NYUD-v2. On PASCAL-Context, it reaches 80.81 mIoU, 73.73 mIoU for human parsing, 86.11 maxF, 13.40 mErr, and 81.18 odsF. These backbone-matched results support the contribution of posterior bridge learning across encoder scales.

\begin{table*}[t]
\centering
\caption{\textcolor{blue}{Efficiency and accuracy comparison on
NYUD-v2. Latency is reported as the mean $\pm$ sample standard
deviation of three session-wise mean latencies. All methods are profiled on the same NVIDIA A100 PCIe 40\,GB GPU with batch size 1.}}
\label{tab:three_methods_complete}

\begingroup
\small
\setlength{\tabcolsep}{4.6pt}
\renewcommand{\arraystretch}{1.10}

\resizebox{\textwidth}{!}{%
\begin{tabular}{@{}llcccccccc@{}}
\toprule
\multirow{3}{*}{Backbone}
& \multirow{3}{*}{Method}
& \multicolumn{4}{c}{Computational cost}
& \multicolumn{4}{c}{Task performance} \\
\cmidrule(lr){3-6}
\cmidrule(lr){7-10}
&
& FLOPs
& Params
& Latency
& Memory
& Semseg
& Depth
& Normal
& Edge \\
&
& (G)$\downarrow$
& (M)$\downarrow$
& Mean $\pm$ SD (ms)$\downarrow$
& (GB)$\downarrow$
& mIoU$\uparrow$
& RMSE$\downarrow$
& mErr$\downarrow$
& odsF$\uparrow$ \\
\midrule

\multirow{3}{*}{ViT-B}
& TaskPrompter
& 1165.8
& 258.5
& \textbf{63.75} $\pm$ 0.05
& 1.401
& 50.40
& 0.5402
& \underline{18.91}
& 77.60 \\

& BridgeNet
& \textbf{741.8}
& \textbf{221.9}
& 82.29 $\pm$ 0.26
& \textbf{1.349}
& \underline{51.14}
& \underline{0.5186}
& 18.92
& \underline{77.98} \\

\rowcolor{oursblue}
& B$^{3}$-Net (ours)
& \underline{745.2}
& \underline{229.0}
& \underline{75.83} $\pm$ 0.03
& \underline{1.377}
& \textbf{52.82}
& \textbf{0.5117}
& \textbf{18.66}
& \textbf{79.24} \\
\midrule

\multirow{3}{*}{ViT-L}
& TaskPrompter
& 1683.8
& 513.5
& \textbf{110.60} $\pm$ 0.19
& 2.446
& 55.30
& 0.5152
& \underline{18.47}
& 78.20 \\

& BridgeNet
& \textbf{1260.6}
& \textbf{463.3}
& 126.20 $\pm$ 0.10
& \textbf{2.303}
& \textbf{55.51}
& \textbf{0.4930}
& \underline{18.47}
& \underline{78.22} \\

\rowcolor{oursblue}
& B$^{3}$-Net (ours)
& \underline{1264.0}
& \underline{470.4}
& \underline{120.57} $\pm$ 0.10
& \underline{2.332}
& \underline{55.46}
& \underline{0.4949}
& \textbf{18.38}
& \textbf{79.75} \\
\midrule

\rowcolor{oursblue}
InternImage-B
& B$^{3}$-Net (ours)
& 701.5
& 243.6
& 79.31 $\pm$ 0.01
& 1.441
& 53.77
& 0.5000
& 18.92
& 82.01 \\

\rowcolor{oursblue}
InternImage-L
& B$^{3}$-Net (ours)
& 918.8
& 390.5
& 88.64 $\pm$ 0.04
& 2.041
& 57.78
& 0.4587
& 17.22
& 83.18 \\

\bottomrule
\end{tabular}%
}
\endgroup
\end{table*}

\textbf{Efficiency and footprint analysis.}
\textcolor{blue}{Table~\ref{tab:three_methods_complete} relates full-model cost to the resulting task trade-off.
Relative to BridgeNet, latency decreases from $82.29$ to $75.83$ ms with ViT-B and from $126.20$ to $120.57$ ms with ViT-L.
Each comparison adds $3.4$ GFLOPs and $7.1$M parameters, with memory increases of $0.028$ and $0.029$ GB, respectively.
ViT-B improves all four tasks, and ViT-L improves normals and edges with close segmentation and depth results.
TaskPrompter has the lowest latency, while BridgeNet has the lowest FLOPs, parameter count, and memory in both ViT groups.
InternImage-L achieves the default four-task results with $918.8$ GFLOPs and $88.64$ ms per image.} Detailed operator-wise footprint and progressive mixed-precision results are provided in Supplementary \textcolor{blue}{Tables~S2 and~S3}.

\textbf{Qualitative comparison.}
Figs.~\ref{fig:qualitative_nyud} and~\ref{fig:qualitative_pascal}
compare B$^{3}$-Net with representative Transformer-, Mamba-,
and bridge-feature-based methods.
On NYUD-v2, competing methods show local semantic leakage
around object intersections and over-smoothed boundaries
near planar structures.
B$^{3}$-Net better preserves the table-screen-chair layout
and produces clearer edge responses.
On PASCAL-Context, B$^{3}$-Net reduces part-level confusion
around the rider-horse contact region, maintains a more
compact foreground response, and gives more continuous
object boundaries.
These qualitative observations are consistent with the
quantitative improvements in segmentation, parsing,
saliency, and edge detection.

Fig.~\ref{fig:output_visualization} visualizes
prediction-derived task maps of B$^{3}$-Net.
The semantic heatmap shows maximum softmax class confidence, the edge map denotes predicted
edge probability, and the last column denotes normalized
depth on NYUD-v2 or predicted saliency on PASCAL-Context.
The maps show that B$^{3}$-Net preserves task-dependent
output structures, including semantic layouts,
object boundaries, indoor geometry, and foreground saliency.

\begin{table*}[t]
\centering
\footnotesize
\setlength{\tabcolsep}{4.2pt}
\renewcommand{\arraystretch}{1.08}
\caption{Operator ablation on NYUD-v2 with B$^{3}$-Net
(Intern-Large).
{\color{newblue}All multi-task variants use seed 42 and
the same training and evaluation settings, differing only
in the enabled PBO, PFE, and CDO operators.
Check marks indicate enabled operators.
The MTL baseline disables all three, while Full enables
all three. The remaining network components are unchanged.}
Task-wise $\Delta_{\mathrm{Task}}(\%)$ and mean $\Delta_{\mathrm{MTL}}(\%)$ gains over STL follow Eqs.~\eqref{eq:deltatau} and~\eqref{eq:deltamtl}.
$\uparrow$/$\downarrow$ denote higher/lower is better.}
\label{tab:ablation_ops_nyud}
\vspace{2pt}
\resizebox{\textwidth}{!}{%
\begin{tabular}{lccc cccc cccc c}
\toprule
\textbf{Variant} &
\textbf{PBO} &
\textbf{PFE} &
\textbf{CDO} &
\textbf{Semseg mIoU}$\uparrow$ &
\textbf{Depth RMSE}$\downarrow$ &
\textbf{Normal mErr}$\downarrow$ &
\textbf{Edge odsF}$\uparrow$ &
\textbf{$\Delta_{\mathrm{Seg}}(\%)\uparrow$} &
\textbf{$\Delta_{\mathrm{Dep}}(\%)\uparrow$} &
\textbf{$\Delta_{\mathrm{Norm}}(\%)\uparrow$} &
\textbf{$\Delta_{\mathrm{Edge}}(\%)\uparrow$} &
\textbf{$\Delta_{\mathrm{MTL}}(\%)\uparrow$} \\
\midrule
STL (per-task)
& -- & -- & --
& 55.65 & 0.4794 & 19.60 & \best{83.19}
& -- & -- & -- & -- & -- \\
MTL baseline
& -- & -- & --
& 53.75 & 0.4805 & 19.96 & 80.60
& -3.41 & -0.23 & -1.84 & -3.11 & -2.15 \\
+ PBO
& \cmark &  &
& 57.50 & \best{0.4570} & 17.41 & 83.18
& 3.32 & \best{4.67} & 11.17 & \best{-0.01} & 4.79 \\
+ PFE
&  & \cmark &
& 57.30 & 0.4588 & 17.45 & 83.01
& 2.96 & 4.30 & 10.97 & -0.22 & 4.50 \\
+ CDO
&  &  & \cmark
& 57.38 & 0.4602 & 17.39 & 82.74
& 3.11 & 4.01 & 11.28 & -0.54 & 4.46 \\
+ PBO + PFE
& \cmark & \cmark &
& 57.63 & 0.4600 & 17.45 & 82.99
& 3.56 & 4.05 & 10.97 & -0.24 & 4.58 \\
+ PBO + CDO
& \cmark &  & \cmark
& 57.52 & 0.4585 & 17.40 & 82.96
& 3.36 & 4.36 & 11.22 & -0.28 & 4.67 \\
+ PFE + CDO
&  & \cmark & \cmark
& 57.60 & 0.4611 & 17.44 & 83.01
& 3.50 & 3.82 & 11.02 & -0.22 & 4.53 \\
\rowcolor{oursblue}
Full: PBO + PFE + CDO (Ours)
& \cmark & \cmark & \cmark
& \best{57.78} & 0.4587 & \best{17.22} & 83.18
& \best{3.83} & 4.32 & \best{12.14} & \best{-0.01} & \best{5.07} \\
\bottomrule
\end{tabular}}
\vspace{-2pt}
\end{table*}

\subsection{Ablation Study}
\textbf{Operator ablation.}
Table~\ref{tab:ablation_ops_nyud} evaluates the contribution of PBO, PFE, and CDO on NYUD-v2.
\textcolor{blue}{The vanilla MTL baseline gives $\Delta_{\mathrm{MTL}}=-2.15\%$, with all four tasks below their STL references.
This motivates examining evidence weighting, bridge formation, and redistribution within the decoder.}

Adding each operator individually changes the transfer behavior from negative to positive.
\textcolor{blue}{PBO, PFE, and CDO yield average gains of $4.79\%$, $4.50\%$, and $4.46\%$, respectively.}
The full model achieves the best overall result, with $\Delta_{\mathrm{MTL}}=5.07\%$.
It also gives the best semantic segmentation and normal estimation gains, with $\Delta_{\mathrm{Seg}}=3.83\%$ and $\Delta_{\mathrm{Norm}}=12.14\%$.
\textcolor{blue}{Removing PBO, PFE, or CDO from Full reduces the average gain by $0.54$, $0.40$, or $0.49$ percentage points.
Pairwise variants remain below PBO alone, showing that the contributions depend on the operator configuration.}

\textcolor{blue}{PBO alone achieves the lowest depth RMSE of $0.4570$, while Full yields $0.4587$.}
The single-task edge baseline obtains 83.19 odsF, while the full model obtains 83.18 odsF.
\textcolor{blue}{Full thus combines the highest average gain with improved segmentation and geometry and an edge score close to STL.
Supplementary Table~S1 reports the matched mean-bridge comparison.}

\textbf{Task-set analysis.}
Tables~\ref{tab:taskset_nyud_full} and~\ref{tab:taskset_pascal_compact} analyze B$^{3}$-Net under different task combinations.
On NYUD-v2, all tested task sets obtain positive average transfer gains.
The Seg+Norm setting achieves the highest two-task gain, with $\Delta_{\mathrm{MTL}}=6.61\%$.
\textcolor{blue}{Full improves all three shared task metrics over Seg+Dep+Norm and raises edge odsF from $82.63$ in Seg+Edge to $83.18$.
Seg+Dep retains higher segmentation mIoU at $59.12$, compared with $57.78$ for Full.
These comparisons show which tasks benefit as the task set changes.}

\begin{table*}[t]
\centering
\scriptsize
\setlength{\tabcolsep}{4.2pt}
\renewcommand{\arraystretch}{1.06}
\caption{Task-set ablation on NYUD-v2 using \textbf{B$^{3}$-Net (Intern-Large)}.}
\label{tab:taskset_nyud_full}
\resizebox{\textwidth}{!}{%
\begin{tabular}{llc cccccccc c}
\toprule
\textbf{Set} & \textbf{Method} & \textbf{$|T|$} &
\textbf{Semseg mIoU}$\uparrow$ &
\textbf{$\Delta_{\mathrm{Seg}}(\%)\uparrow$} &
\textbf{Depth RMSE}$\downarrow$ &
\textbf{$\Delta_{\mathrm{Dep}}(\%)\uparrow$} &
\textbf{Normal mErr}$\downarrow$ &
\textbf{$\Delta_{\mathrm{Norm}}(\%)\uparrow$} &
\textbf{Edge odsF}$\uparrow$ &
\textbf{$\Delta_{\mathrm{Edge}}(\%)\uparrow$} &
\textbf{$\Delta_{\mathrm{MTL}}(\%)\uparrow$} \\
\midrule
-- & STL (per-task) & -- &
55.65 & -- &
0.4794 & -- &
19.60 & -- &
83.19 & -- &
-- \\
\midrule
Seg+Dep & B$^{3}$-Net (Intern-L) & 2
& 59.12 & 6.24
& 0.4580 & 4.46
& \multicolumn{2}{c}{--}
& \multicolumn{2}{c}{--}
& 5.35 \\
Seg+Norm & B$^{3}$-Net (Intern-L) & 2
& 56.85 & 2.16
& \multicolumn{2}{c}{--}
& 17.43 & 11.07
& \multicolumn{2}{c}{--}
& 6.61 \\
Seg+Edge & B$^{3}$-Net (Intern-L) & 2
& 58.94 & 5.91
& \multicolumn{2}{c}{--}
& \multicolumn{2}{c}{--}
& 82.63 & -0.67
& 2.62 \\
Seg+Dep+Norm & B$^{3}$-Net (Intern-L) & 3
& 57.45 & 3.23
& 0.4611 & 3.82
& 17.43 & 11.07
& \multicolumn{2}{c}{--}
& 6.04 \\
Full (4 tasks) & B$^{3}$-Net (Intern-L) & 4
& 57.78 & 3.83
& 0.4587 & 4.32
& 17.22 & 12.14
& 83.18 & -0.01
& 5.07 \\
\bottomrule
\end{tabular}}
\end{table*}

The vanilla MTL baseline obtains $\Delta_{\mathrm{MTL}}=-3.35\%$ under five-task learning.
\textcolor{blue}{On the same PASCAL-Context task set, Full reaches $3.22\%$.}
It improves all five tasks over the single-task baselines.
\textcolor{blue}{Omitting saliency raises parsing mIoU from $73.73$ to $73.95$ and edge odsF from $81.18$ to $81.85$, while normal error remains $13.40$.
Since averaged gains involve different task sets, these common-task comparisons identify the specific trade-offs.
The varying responses motivate local evidence weighting and task-specific adaptation within the shared bridge.}

\begin{table*}[t]
\centering
\scriptsize
\setlength{\tabcolsep}{4.6pt}
\renewcommand{\arraystretch}{1.08}
\caption{Comparison of different task sets on PASCAL-Context using \textbf{B$^{3}$-Net (Intern-Large)}. }
\label{tab:taskset_pascal_compact}
\vspace{2pt}
\resizebox{\textwidth}{!}{%
\begin{tabular}{l cc cc cc cc cc c}
\toprule
\textbf{Method} &
\textbf{Semseg mIoU}$\uparrow$ & \textbf{$\Delta_{\mathrm{Seg}}(\%)\uparrow$} &
\textbf{Parsing mIoU}$\uparrow$ & \textbf{$\Delta_{\mathrm{Par}}(\%)\uparrow$} &
\textbf{Saliency maxF}$\uparrow$ & \textbf{$\Delta_{\mathrm{Sal}}(\%)\uparrow$} &
\textbf{Normal mErr}$\downarrow$ & \textbf{$\Delta_{\mathrm{Norm}}(\%)\uparrow$} &
\textbf{Edge odsF}$\uparrow$ & \textbf{$\Delta_{\mathrm{Edge}}(\%)\uparrow$} &
\textbf{$\Delta_{\mathrm{MTL}}(\%)\uparrow$} \\
\midrule
STL (per-task)
& 80.00 & --  & 71.40 & --  & 85.36 & --  & 15.00 & --  & 80.97 & --  & -- \\
MTL baseline
& 76.33 & -4.59
& 66.21 & -7.27
& 84.07 & -1.51
& 15.42 & -2.80
& 80.49 & -0.59
& -3.35 \\
\midrule
Full (Seg+Par+Sal+Norm+Edge)
& 80.81 & 1.01
& 73.73 & 3.26
& 86.11 & 0.88
& 13.40 & 10.67
& 81.18 & 0.26
& 3.22 \\
Seg+Sal+Norm+Edge
& 80.10 & 0.13
& -- & --
& 86.03 & 0.78
& 13.19 & 12.07
& 81.06 & 0.11
& 3.27 \\
Seg+Par+Norm+Edge
& 80.78 & 0.98
& 73.95 & 3.57
& -- & --
& 13.40 & 10.67
& 81.85 & 1.09
& 4.08 \\
Seg+Par+Sal+Edge
& 82.49 & 3.11
& 75.90 & 6.30
& 85.86 & 0.59
& -- & --
& 82.91 & 2.40
& 3.10 \\
\bottomrule
\end{tabular}}
\vspace{-2pt}
\end{table*}

\begin{figure}[!t]
\centering
\includegraphics[width=0.85\columnwidth]{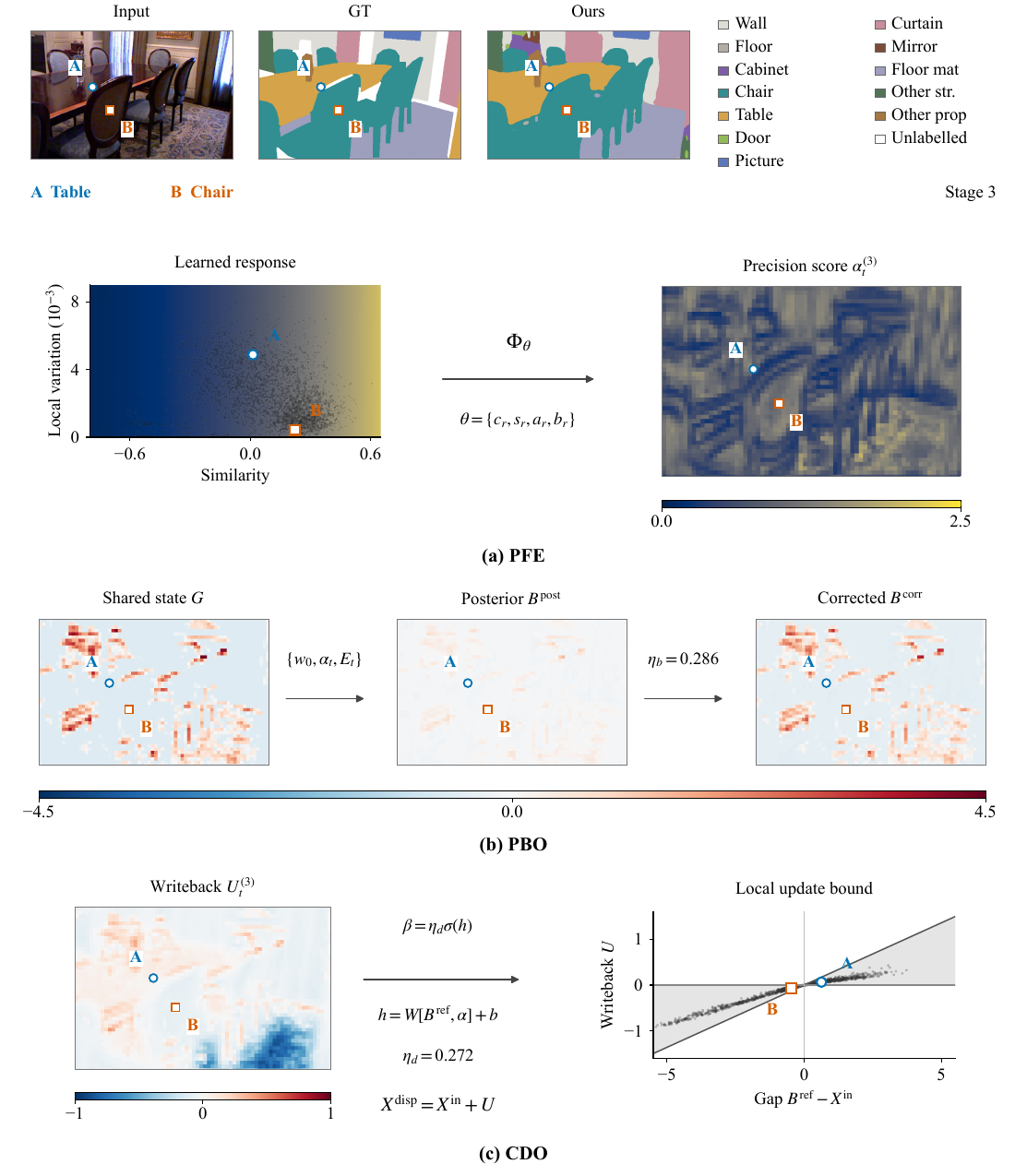}
\caption{\textcolor{blue}{\textbf{Stage-3 operator mechanisms on NYUD-v2.}
(a) Learned PFE response and spatial precision.
(b) Posterior formation and pre-MLP correction.
(c) CDO writeback and componentwise bound.}}
\label{fig:spatial-operator-mechanism}
\end{figure}

\textbf{Mechanism verification.}
{\color{blue}
Fig.~\ref{fig:spatial-operator-mechanism} traces the three operators on the same NYUD-v2 image at Stage~3.
A and B mark table and chair locations in the input, GT, predictions, and internal maps.
In (a), learned centers $c_r$ and scales $s_r$ determine rule activation, while $(a_r,b_r)$ determine the rule outputs.
The response surface maps the distinct similarity and variation at A and B to different spatial precision scores.
The stage-level rule bank is fixed at inference, giving a shared mapping with location-dependent weights.
\par}

In PBO, the precision value directly controls the contribution of task evidence to the shared bridge.
\textcolor{blue}{Panel (b) shows the resulting $B^{\mathrm{post}}$ and its pre-MLP correction $B^{\mathrm{corr}}$, corresponding to $B$ and $\widehat B$.
On the common color scale, the correction retains much of $G$'s spatial pattern because $\eta_b\simeq0.286$ adopts part of the posterior displacement.
The Gaussian working model acts on continuous latent evidence without a spatial smoothing penalty.
Positive precisions yield a unique quadratic minimizer.
Its conditional minimum-variance interpretation assumes zero-mean, mutually uncorrelated isotropic source errors and precisions proportional to their inverse variances.}

\textcolor{blue}{Panel (c) pairs each cell's gap with its actual writeback,
$U_c=\beta_c(B_c^{\mathrm{ref}}-X_c^{\mathrm{in}})$.
The learned gate $\beta=\eta_{\mathrm d}\sigma(W[B^{\mathrm{ref}},\alpha]+b)$ controls this update, with $\eta_{\mathrm d}\simeq0.272$.
The points lie between zero and the line of slope $\eta_{\mathrm d}$, showing gap-aligned updates of bounded magnitude.
Each updated component remains between its input and fixed reference within this single dispatch, before subsequent refinement.}
It controls how much shared information each task receives after posterior bridge formation.
\textcolor{blue}{Supplementary Figs.~S1-S3 provide the precision--error analysis, CDO statistics, and posterior objective slice.}

{\color{blue}
\textbf{Posterior correction sensitivity.}
Table~\ref{tab:eta_b_sensitivity} tests how much of the estimated posterior should enter the corrected reference.
The endpoints $\eta_b=0$ and $1$ retain $G$ and adopt the full posterior, respectively.
Average gain increases from $4.52\%$ to $4.75\%$ and $4.90\%$ across  $0$, $0.5$, and $1$.
Stage-wise learning reaches $5.07\%$ and leads segmentation, depth, and normals, while $\eta_b=1$ gives the highest edge odsF.
The $0.17$-point improvement over full posterior adoption supports learning the correction strength in addition to estimating the posterior.
\par}

\begin{table}[t]
\centering
\caption{\textcolor{blue}{Sensitivity to the posterior-correction coefficient
$\eta_b$ on NYUD-v2 with InternImage-Large.
Values $0$ and $1$ denote the limiting cases.}}
\label{tab:eta_b_sensitivity}
\setlength{\tabcolsep}{4.6pt}
\renewcommand{\arraystretch}{1.10}
\resizebox{\columnwidth}{!}{
\begin{tabular}{@{}lccccc@{}}
\toprule
Setting
& Semseg
& Depth
& Normal
& Edge
& $\Delta_{\mathrm{MTL}}$ \\
& mIoU$\uparrow$
& RMSE$\downarrow$
& mErr$\downarrow$
& odsF$\uparrow$
& (\%)$\uparrow$ \\
\midrule
STL (per-task)
& 55.65
& 0.4794
& 19.60
& 83.19
& -- \\
\midrule
$\eta_b=0$
& 57.09
& 0.4642
& 17.25
& \underline{83.49}
& 4.52 \\

$\eta_b=0.5$
& 57.31
& \underline{0.4607}
& 17.28
& 83.40
& 4.75 \\

$\eta_b=1$
& \underline{57.47}
& 0.4613
& \underline{17.23}
& \textbf{83.58}
& \underline{4.90} \\

\rowcolor{oursblue}
Learned (Ours)
& \textbf{57.78}
& \textbf{0.4587}
& \textbf{17.22}
& 83.18
& \textbf{5.07} \\
\bottomrule
\end{tabular}
}
\end{table}

\begin{table}[t]
\centering
\caption{\textcolor{blue}{Sensitivity to PFE spatial granularity
on NYUD-v2 with InternImage-Large.
STL provides the per-task references for computing
$\Delta_{\mathrm{MTL}}$.
The default $1\times1$ setting reuses the main-experiment result.
The $2\times2$ and $4\times4$ settings are independently trained
under the same protocol.}}
\label{tab:pfe-granularity}
\setlength{\tabcolsep}{4.6pt}
\renewcommand{\arraystretch}{1.10}
\resizebox{\columnwidth}{!}{
\begin{tabular}{@{}lccccc@{}}
\toprule
PFE block
& Semseg
& Depth
& Normal
& Edge
& $\Delta_{\mathrm{MTL}}$ \\
& mIoU$\uparrow$
& RMSE$\downarrow$
& mErr$\downarrow$
& odsF$\uparrow$
& (\%)$\uparrow$ \\
\midrule
STL (per-task)
& 55.65
& 0.4794
& 19.60
& 83.19
& -- \\
\midrule
$4\times4$
& \underline{57.14}
& \underline{0.4622}
& 17.27
& 83.09
& 4.50 \\

$2\times2$
& 57.09
& \underline{0.4622}
& \underline{17.26}
& \underline{83.13}
& \underline{4.51} \\

\rowcolor{oursblue}
$1\times1$ (Ours)
& \textbf{57.78}
& \textbf{0.4587}
& \textbf{17.22}
& \textbf{83.18}
& \textbf{5.07} \\
\bottomrule
\end{tabular}
}
\end{table}

\begin{figure}[!t]
\centering
\includegraphics[width=0.85\columnwidth]{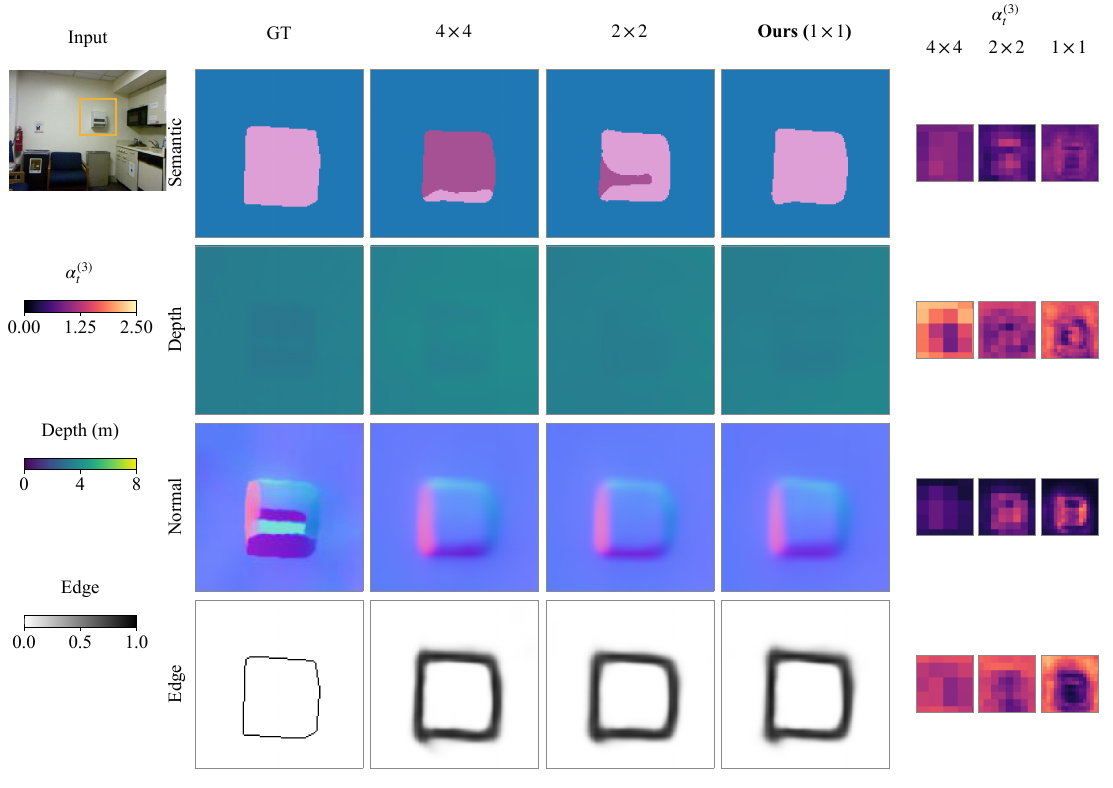}
\caption{\textcolor{blue}{\textbf{PFE granularity and local predictions on NYUD-v2.}
All tasks share the marked crop.
The right-hand maps show actual Stage-3 precision on the common scale $[0,2.5]$.}}
\label{fig:pfe_joint_granularity}
\end{figure}

{\color{blue}
\textbf{PFE spatial granularity.}
Table~\ref{tab:pfe-granularity} examines the spatial resolution of evidence weighting.
The default $1\times1$ setting assigns one precision per native cell, giving $56\times72$ locations at Stage~3.
It leads all four metrics, exceeding $2\times2$ and $4\times4$ by $0.56$ and $0.57$ percentage points in average gain.
Fig.~\ref{fig:pfe_joint_granularity} shows the corresponding task predictions and precision maps for the same crop.
Coarse precision maps share weights over larger blocks, and both coarse models introduce an interior semantic class error.
The default resolves finer precision variation and preserves the GT class within the object.
Depth varies little in this crop, and normal and edge predictions have similar overall structure.
The local semantic difference and full-validation metrics jointly support native-cell precision for spatially selective fusion.
\par}

\textcolor{blue}{\textbf{PFE parameterization.}}
Table~\ref{tab:pfe_design_nyud_formal} further compares different PFE parameterizations under clean and perturbed inputs. {\color{newblue}For Table~\ref{tab:pfe_design_nyud_formal}, perturbations are applied only during evaluation on the NYUD-v2 validation split.
After ImageNet standardization with channel means $(0.485,0.456,0.406)$ and standard deviations $(0.229,0.224,0.225)$, each image receives independent Gaussian noise with standard deviation $0.03$ at each pixel and channel, followed by a channel-wise $3\times3$ box filter with weights $1/9$, stride 1, and zero padding of 1.
Perturbation parameters are fixed across models, while noise is sampled online for each validation image.
The resulting tensors are passed directly to the network without clipping or further normalization.
Each PFE variant uses the same trained checkpoint for its clean and perturbed evaluations.
\par}
PFE-Fuzzy obtains the best edge performance under both clean and noisy settings and gives the best normal result under noise.
PFE-CNN gives the lowest depth RMSE, while PFE-Const performs well on semantic segmentation.
\textcolor{blue}{The rule mapping in Fig.~\ref{fig:spatial-operator-mechanism}(a) is thus supported by its edge and perturbed-normal results in this comparison, complementing the spatial-resolution analysis above.}

\begin{table}[t]
\centering
\scriptsize
\setlength{\tabcolsep}{3.0pt}
\renewcommand{\arraystretch}{1.12}
\caption{\textbf{Formal comparison of PFE variants on NYUD-v2 with B$^{3}$-Net (Intern-Large).}
Each variant is trained independently under matched optimization settings and training budgets, with only the PFE parameterization changed. ``Noise'' denotes evaluation under a fixed perturbation protocol (Gaussian noise + mild blur). Edge performance is reported using externally evaluated ODS F-measure.}
\label{tab:pfe_design_nyud_formal}

\resizebox{\columnwidth}{!}{%
\begin{tabular}{@{}lccccc@{}}
\toprule
\textbf{Method}
& \shortstack{\textbf{Semseg}\\mIoU$\uparrow$}
& \shortstack{\textbf{Depth}\\RMSE$\downarrow$}
& \shortstack{\textbf{Normal}\\mErr$\downarrow$}
& \shortstack{\textbf{Edge}\\odsF$\uparrow$}
& \shortstack{\textcolor{newblue}{$\Delta_{\mathrm{MTL}}$}\\
              \textcolor{newblue}{(\%)$\uparrow$}} \\
\midrule

\textcolor{newblue}{STL}
& \textcolor{newblue}{55.65}
& \textcolor{newblue}{0.4794}
& \textcolor{newblue}{19.60}
& \textcolor{newblue}{83.19}
& \textcolor{newblue}{--} \\
\midrule

PFE-Const (clean)
& \best{57.52}
& 0.4629
& \second{17.423}
& \second{82.96}
& \textcolor{newblue}{4.41} \\

PFE-Const + Noise
& \best{56.02}
& 0.5102
& \second{18.046}
& \second{81.16}
& \textcolor{newblue}{$-0.07$} \\

PFE-CNN (clean)
& \second{57.49}
& \best{0.4604}
& \best{17.420}
& 82.19
& \textcolor{newblue}{4.30} \\

PFE-CNN + Noise
& \second{55.99}
& \best{0.4990}
& 18.097
& 80.21
& \textcolor{newblue}{0.15} \\

PFE-Fuzzy (clean)
& 57.19
& \second{0.4619}
& 17.447
& \best{83.03}
& \textcolor{newblue}{4.30} \\

PFE-Fuzzy + Noise
& 55.72
& \second{0.5044}
& \best{18.032}
& \best{81.23}
& \textcolor{newblue}{0.14} \\
\bottomrule
\end{tabular}%
}
\end{table}

\begin{figure}[!t]
\centering
\includegraphics[width=0.85\columnwidth]{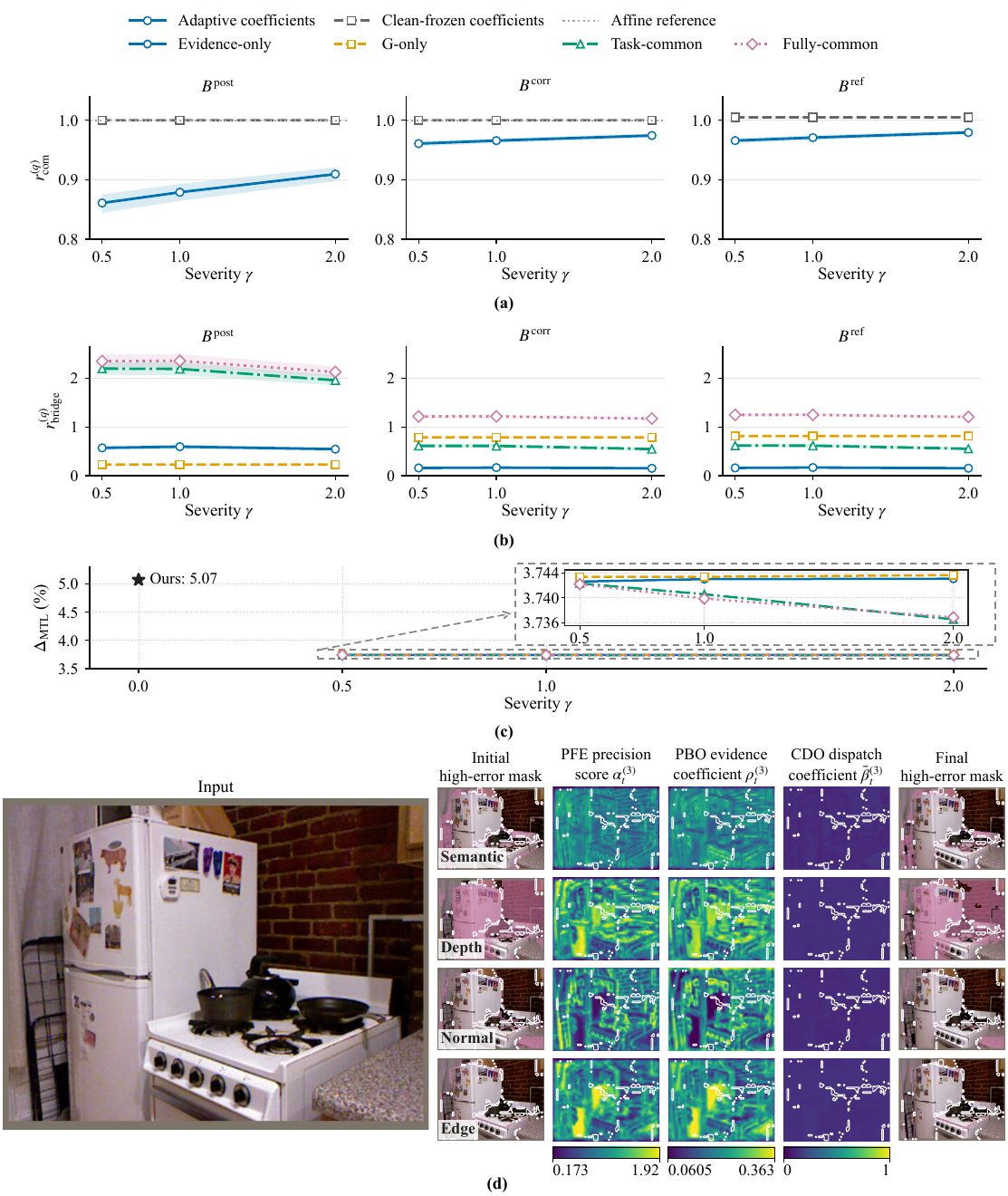}
\caption{\textcolor{blue}{\textbf{Common-bias responses and spatial co-failure.}
(a) Common-direction response. The unit affine reference applies to frozen posterior and correction states with zero stabilizer.
(b) Bridge response across perturbation scopes.
Shading denotes 95\% image-level bootstrap intervals over 40 images with 2,000 draws after within-image aggregation.
(c) Task gain on all 654 validation images with one fixed task-stage-direction assignment per image and full-field perturbation.
(d) Initial/final masks use identical validity masks and training-split 90th-percentile loss thresholds.
Magenta marks high error, and white outlines initial overlap of at least three tasks.
Stage-3 maps show precision, evidence coefficients, and channel-averaged dispatch.
The case follows the median-overlap rule.}}
\label{fig:common-bias-cofailure}
\end{figure}

{\color{blue}
\textbf{Shared-reference sensitivity and common errors.}
Fig.~\ref{fig:common-bias-cofailure} examines where the same interaction remains sensitive to perturbed information.
It tests one evidence source, $G$ alone, all task evidence, or both $G$ and all evidence.
For state $q\in\{\mathrm{post},\mathrm{corr},\mathrm{ref}\}$ and injected field $\delta(x,c)=M(x)b(c)$,
(a) measures the response along the injected direction and (b) its magnitude,
\[
\begin{aligned}
r_{\mathrm{com}}^{(q)}
&=\frac{\sum_{x,c}(B'^{(q)}-B^{(q)})\delta}{\sum_{x,c}\delta^2},\\
r_{\mathrm{bridge}}^{(q)}
&=\frac{1}{|\Omega_M|}\sum_{x\in\Omega_M}
\frac{\|B'^{(q)}(x,:)-B^{(q)}(x,:)\|_2}{\|b\|_2}.
\end{aligned}
\]
Here $\Omega_M$ contains the injected locations.
Normalization is per source, so the scopes have different total perturbation energies.
Adaptive coefficients in (a) lower the posterior response below the frozen reference of one, while the corrected and refined responses remain near one.
In (b), task-common perturbations affect the posterior more than $G$-only perturbations, and this ordering reverses after correction.
For fixed normalized coefficients, $G$ retains coefficient $1-\eta_b+\eta_b\rho_0$, where $\rho_0$ is its posterior weight.
Partial correction therefore preserves reference information together with reference error.

In (c), task-common and fully-common curves decrease slightly with severity, while evidence-only and $G$-only curves vary little around the internal control of $3.7426\%$.
The $5.07\%$ star is the independent main-experiment result.
Panel (d) localizes persistent errors in the same scene through initial and final masks.
Magenta regions remain within the white initial-overlap contours despite nonzero precision and evidence coefficients and bounded dispatch.
This case connects simultaneous prediction errors to a practical boundary of controlled sharing, where local update bounds coexist with unresolved errors across tasks.
\par}

\section{Conclusion}
$\mathcal{B}^{3}$-Net addresses decoder-side negative transfer
through patch-wise PFE precision, precision-weighted PBO fusion,
and reference-relative single-dispatch CDO bounds.
Gaussian fusion has a unique closed-form solution for fixed
evidence, reference, and positive precisions.
Its spatially varying latent means and precisions accommodate
structural changes through a spatially separable objective.
Under Sec.~III-B's conditional error and inverse-variance
calibration assumptions, exact normalized PBO minimizes variance
among scalar-weighted unbiased combinations.
The supplement analyzes Laplace and Student-$t$ alternatives.
NYUD-v2, PASCAL-Context, and Cityscapes evaluations show strong
reported-setting and backbone-matched performance across
heterogeneous task sets with lightweight operators.
Operator, correction, granularity, and mean-bridge ablations
support precision-guided aggregation and bounded adaptation,
with aggregate depth precision--error alignment.

\textcolor{newblue}{Our training runs use seed 42.
On two-task Cityscapes, BridgeNet and $\mathcal{B}^{3}$-Net
outperform the remaining methods across all metrics.
Our model gains 0.23 mIoU points over BridgeNet and reduces
Abs. Rel. and Sq. Rel. by 0.0010 and 0.008.
When reference and task evidence are close, normalized
reweighting yields small bridge changes.
These margins and CDO's bounded updates motivate selective
refinement according to local evidence disagreement.}
Shared representation errors or missing information can bias
$G$ and all task evidence, leaving a common error in exact
normalized fusion.
Spatial co-failures motivate precision calibration using
task-dependent residual statistics and cross-task error dependence,
and reference estimation from complementary, aligned temporal
or multimodal observations.
Separating source reliability from receiver-specific updates
through an explicit shared state makes evidence selection,
fusion, and use analyzable, potentially guiding multimodal
perception, video understanding, and multi-view reconstruction.


 



\vspace{-22pt}
\begin{IEEEbiography}[{\includegraphics[width=1in,height=1.25in,clip,keepaspectratio]{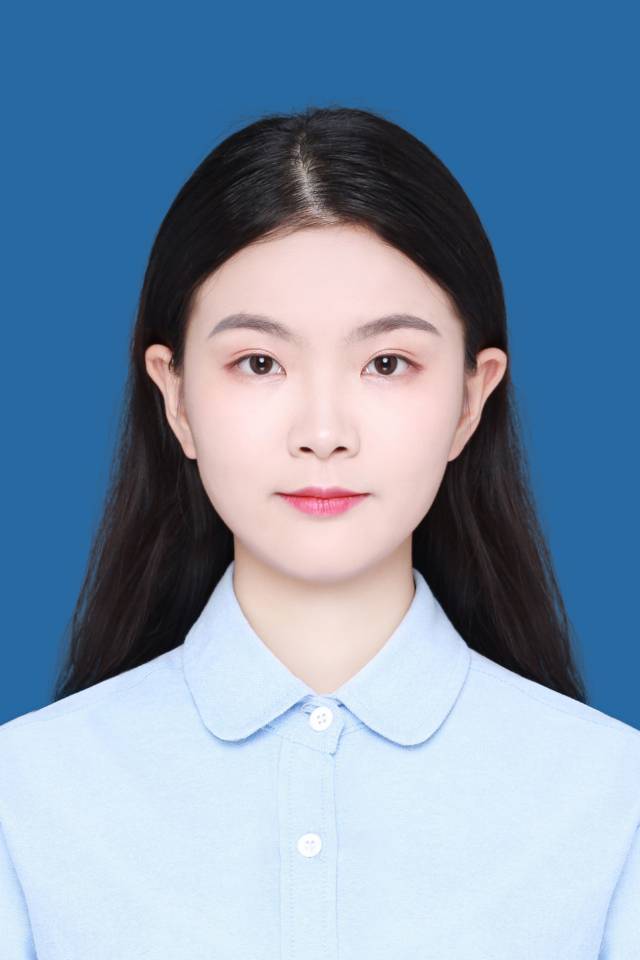}}]{Meihua Zhou}
received the B.S. degree in Information Management and Information Systems from Wannan Medical University, Wuhu, China, in 2023. From 2024 to 2025, she worked on multi-task learning and multimodal visual computing in the Department of Computer Science, Tsinghua University, Beijing, China, while also acquiring interdisciplinary research experience at the Ophthalmology Research Institute of Beijing Tongren Hospital, Beijing, China. She is currently pursuing graduate study at the University of Chinese Academy of Sciences, Beijing, China. Her research interests include multi-task learning, dense visual prediction, multimodal computing, embodied intelligence, reliable representation learning, and interdisciplinary applications of artificial intelligence in medical and real-world intelligent systems.
\end{IEEEbiography}

\begin{IEEEbiography}[{\includegraphics[width=1in,height=1.25in,clip,keepaspectratio]{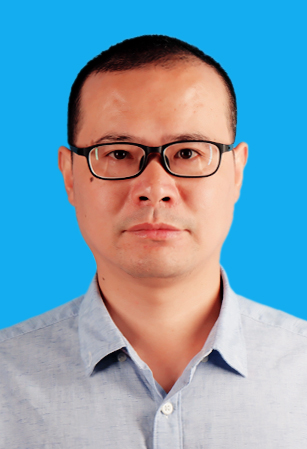}}]{Li Yang}
received the B.E. degree in Computer Science and Technology from Anhui University, Hefei, China, in 2006, and the M.E. degree in Computer Technology from Anhui University, Hefei, China, in 2012. In 2019, he participated in an exchange program at the University of Hong Kong, Hong Kong, China, as a young backbone teacher of Wannan Medical University. From 2023 to 2024, he was a visiting scholar at the University of Science and Technology of China, Hefei, China, where he focused on artificial intelligence and its interdisciplinary applications. He is currently an Associate Professor with Wannan Medical University, Wuhu, China. His research interests include artificial intelligence, computer vision, multi-task learning.
\end{IEEEbiography}

\vfill

\end{document}